\documentclass{article} %
\usepackage{iclr2021_conference,times}

\usepackage{amsmath,amsfonts,bm}

\def\eqref#1{equation~\ref{#1}}

\def\1{\bm{1}}

\def\vb{{\bm{b}}}

\def\vg{{\bm{g}}}

\def\vn{{\bm{n}}}

\def\vt{{\bm{t}}}

\def\vx{{\bm{x}}}

\def\evx{{x}}

\def\mA{{\bm{A}}}

\def\mC{{\bm{C}}}

\def\mH{{\bm{H}}}
\def\mI{{\bm{I}}}
\def\mJ{{\bm{J}}}

\def\mL{{\bm{L}}}
\def\mM{{\bm{M}}}

\def\mQ{{\bm{Q}}}

\def\mT{{\bm{T}}}
\def\mU{{\bm{U}}}

\def\mW{{\bm{W}}}

\DeclareMathAlphabet{\mathsfit}{\encodingdefault}{\sfdefault}{m}{sl}
\SetMathAlphabet{\mathsfit}{bold}{\encodingdefault}{\sfdefault}{bx}{n}
\newcommand{\tens}[1]{\bm{\mathsfit{#1}}}

\def\tD{{\tens{D}}}

\def\tF{{\tens{F}}}
\def\tG{{\tens{G}}}
\def\tH{{\tens{H}}}

\def\emA{{A}}

\def\emH{{H}}

\def\emU{{U}}

\newcommand{\etens}[1]{\mathsfit{#1}}

\def\etD{{\etens{D}}}

\def\etG{{\etens{G}}}
\def\etH{{\etens{H}}}

\newcommand{\E}{\mathbb{E}}

\newcommand{\R}{\mathbb{R}}

\usepackage{amsmath, amssymb}
\usepackage{amsthm}
\usepackage{bm}

\usepackage{hhline}

\usepackage{hyperref}
\usepackage{url}

\usepackage{makecell}
\usepackage{float}
\usepackage{caption}
\usepackage{graphicx}
\usepackage{stackengine}
\usepackage{textcomp}
\usepackage{multirow}
\usepackage{wrapfig}

\title{Isometric Transformation Invariant and Equivariant Graph Convolutional Networks}

\author{
Masanobu Horie\\
University of Tsukuba,\\
Research Institute for Computational Science Co. Ltd.\\
\texttt{horie@ricos.co.jp} \\
\And
Naoki Morita\\
University of Tsukuba,\\
Research Institute for Computational Science Co. Ltd.\\
\texttt{morita@ricos.co.jp} \\
\AND
Toshiaki Hishinuma \& Yu Ihara\\
Research Institute for Computational Science Co. Ltd.\\
\texttt{\{hishinuma,ihara\}@ricos.co.jp} \\
\And
Naoto Mitsume\\
University of Tsukuba\\
\texttt{mitsume@kz.tsukuba.ac.jp}\\
}

\newcommand{\sub}[1]{_\mathrm{#1}}

\newcommand{\supp}[1]{^{(#1)}}
\newcommand{\G}{\mathcal{G}}
\newcommand{\V}{\mathcal{V}}
\newcommand{\VV}{\vert\mathcal{V}\vert}
\renewcommand{\E}{\mathcal{E}}

\newcommand{\Z}{\mathbb{Z}}

\renewcommand{\sup}[1]{^\mathrm{#1}}
\renewcommand{\O}{\mathrm{O}}
\newcommand{\pdiff}[2]{\frac{\partial #1}{\partial #2}}
\newcommand{\inpdiff}[2]{\partial #1 / \partial #2}

\newtheorem{proposition}{Proposition}[section]

\iclrfinalcopy %
\begin{document}

\maketitle

\begin{abstract}
  Graphs are one of the most important data structures for representing pairwise relations between objects.
  Specifically, a graph embedded in a Euclidean space is essential to solving
  real problems, such as physical simulations.
  A crucial requirement for applying graphs in Euclidean spaces to physical simulations is
  learning and inferring the isometric transformation invariant and equivariant features
  in a computationally efficient manner.
  In this paper, we propose a set of transformation invariant and equivariant
  models based on graph convolutional networks, called IsoGCNs.
  We demonstrate that the proposed model has a competitive performance compared to state-of-the-art
  methods on tasks related to geometrical and physical simulation data.
  Moreover, the proposed model can scale up to graphs with 1M vertices
  and conduct an inference faster than a conventional finite element analysis,
  which the existing equivariant models cannot achieve.
\end{abstract}

\section{Introduction}
Graph-structured data embedded in Euclidean spaces can be utilized in
many different fields such as object detection, structural chemistry analysis, and physical
simulations.
Graph neural networks (GNNs) have been introduced to deal with such data.
The crucial properties of GNNs include permutation invariance and equivariance.
Besides permutations, isometric transformation invariance and
equivariance must be addressed when considering graphs in Euclidean spaces
because many properties of objects in the Euclidean space do not change
under translation and rotation.
Due to such invariance and equivariance,
1) the interpretation of the model is facilitated;
2) the output of the model is stabilized and predictable;
and 3) the training is rendered efficient by eliminating the necessity of data augmentation
as discussed in the literature~\citep{thomas2018tensor,weiler20183d,fuchs2020se}.

Isometric transformation invariance and equivariance are inevitable,
especially when applied to physical simulations,
because every physical quantity and physical law is either invariant or equivariant to such a transformation.
Another essential requirement for such applications is computational efficiency
because the primary objective of learning a physical simulation is to replace
a computationally expensive simulation method with a faster machine learning model.

In the present paper, we propose \emph{IsoGCNs}, a set of simple yet powerful models that provide
computationally-efficient isometric transformation invariance and equivariance based on
graph convolutional networks (GCNs)~\citep{kipf2017semi}.
Specifically, by simply tweaking the definition of an adjacency matrix,
the proposed model can realize isometric transformation
invariance.
Because the proposed approach relies on graphs, it can deal with the complex shapes that are usually
presented using mesh or point cloud data structures.
Besides, a specific form of the IsoGCN layer can be regarded as a spatial
differential operator that is essential for describing physical laws.
In addition, we have shown that the proposed approach is computationally efficient in terms of processing graphs with up to
1M vertices that are often presented in real physical simulations.
Moreover, the proposed model exhibited faster inference compared to
a conventional finite element analysis approach at the same level of accuracy.
Therefore, an IsoGCN can suitably replace physical simulations regarding its
power to express physical laws and faster, scalable computation.
The corresponding implementation and the dataset are available online\footnote{\url{https://github.com/yellowshippo/isogcn-iclr2021}}.

The main contributions of the present paper can be summarized as follows:
\begin{itemize}
  \item
    We construct isometric invariant and equivariant GCNs, called IsoGCNs
    for the specified input and output tensor ranks.
  \item
    We demonstrate that an IsoGCN model enjoys competitive performance against state-of-the-art baseline models
    on the considered tasks related to physical simulations.
  \item
    We confirm that IsoGCNs are scalable to
    graphs with 1M vertices and achieve inference considerably faster than
    conventional finite element analysis.
\end{itemize}

\section{Related work}
\textbf{Graph neural networks.}
The concept of a GNN was first proposed by~\citet{baskin1997neural, sperduti1997supervised}
and then improved by~\citep{gori2005new, scarselli2008graph}.
Although many variants of GNNs have been proposed,
these models have been unified under the concept of message passing neural networks~\citep{gilmer2017neural}.
Generally, message passing is computed with nonlinear neural networks, which can incur a tremendous
computational cost.
In contrast, the GCN developed by~\citet{kipf2017semi} is a considerable simplification of a GNN,
that uses a linear message passing scheme expressed as
\begin{align}
  \mH\sub{out} = \sigma(\hat{\mA} \mH\sub{in} \mW),
  \label{eq:gcn}
\end{align}
where $\mH\sub{in}$ ($\mH\sub{out}$) is an input (output) feature of the $l$th layer, $\hat{\mA}$ is a renormalized
adjacency matrix with self-loops, and $\mW$ is a trainable weight.
A GCN, among the variants of GNNs, is essential to the present study
because the proposed model is based on GCNs for computational efficiency.

\textbf{Invariant and equivariant neural networks.}
A function $f: X \rightarrow Y$ is said to be equivariant to a group $G$ when $f(g \cdot x) = g \cdot f(x)$,
for all $g \in G$ and $x \in X$, assuming that group $G$ acts on both $X$ and $Y$.
In particular, when $f(g \cdot x) = f(x)$, $f$ is said to be invariant to the group $G$.
Group equivariant convolutional neural networks were first proposed by \citet{cohen2016group} for discrete groups.
Subsequent studies have categorized such networks into continuous groups~\citep{cohen2018spherical},
three-dimensional data~\citep{weiler20183d}, and general manifolds~\citep{cohen2019gauge}.
These methods are based on CNNs; thus, they cannot handle mesh or point cloud data structures as is.
Specifically, 3D steerable CNNs~\citep{weiler20183d} uses voxels (regular grids),
which though relatively easy to handle, are
not efficient because they represent both occupied and non-occupied parts of an object~\citep{ahmed2018survey}.
In addition, a voxelized object tends to lose the smoothness of its shape, which can lead to drastically different
behavior in a physical simulation, as typically observed in structural analysis and computational fluid dynamics.

\citet{thomas2018tensor, kondor2018n} discussed how to provide rotation equivariance to point clouds.
Specifically, the tensor field network (TFN)~\citep{thomas2018tensor} is a point cloud based rotation
and translation equivariant neural network the layer of which can be written as
\begin{align}
  \tilde{\tH}_{\mathrm{out}, i}\supp{l} &= w^{ll} \ \tilde{\tH}_{\mathrm{in}, i}\supp{l}
  + \sum_{k \geq 0} \sum_{j \neq i} \mW^{lk}(\vx_j - \vx_i) \ \tilde{\tH}_{\mathrm{in}, j}\supp{k},
  \label{eq:tfn}
  \\
  \mW^{lk}(\vx) &= \sum_{J = \vert k - l \vert}^{k + l} \phi_J^{lk}(\Vert \vx \Vert)
  \sum_{m=-J}^{J} Y_{Jm}(\vx / \Vert \vx \Vert) \mQ^{lk}_{Jm},
\end{align}
where $\tilde{\tH}_{\mathrm{in}, i}\supp{l}$ ($\tilde{\tH}_{\mathrm{out}, i}\supp{l}$) is a
type-$l$ input (output) feature at the $i$th vertex,
$\phi_J^{lk}: \mathbb{R}_{\geq0} \to \mathbb{R}$ is a trainable function,
$Y_{Jm}$ is the $m$th component of the $J$th spherical harmonics, and
$\mQ^{lk}_{Jm}$ is the Clebsch-Cordan coefficient.
The SE(3)-Transformer~\citep{fuchs2020se} is a variant of the TFN with self-attention.
These models achieve high expressibility based on spherical harmonics and message passing with nonlinear neural networks.
However, for this reason, considerable computational resources are required.
In contrast, the present study allows a significant reduction in the computational costs
because it eliminates spherical harmonics and nonlinear message passing.
From this perspective, IsoGCNs are also regarded as a simplification of the TFN,
as seen in equation~\ref{eq:isogcn}.

\textbf{Physical simulations using GNNs.}
Several related studies, including those by \citet{sanchez2018graph, sanchez2019hamiltonian, alet19a, chang2020learning}
focused on applying GNNs to learn physical simulations.
These approaches allowed the physical information to be introduced to GNNs; however,
addressing isometric transformation equivariance was out of the scope of their research.

In the present study, we incorporate isometric transformation invariance and equivariance into GCNs,
thereby, ensuring the stability of the training and inference under isometric transformation.
Moreover, the proposed approach is efficient in processing large graphs with up to 1M vertices
that have a sufficient number of degrees of freedom to express complex shapes.

\section{Isometric Transformation invariant and equivariant graph convolutional layers}\label{sec:model}
In this section, we discuss how to construct IsoGCN layers that
correspond to the isometric invariant and equivariant GCN layers.
To formulate a model, we assume that:
1) only attributes associated with vertices and not edges; and
2) graphs do not contain self-loops.
Here, $\G = (\V, \E)$ denotes a graph and $d$ denotes the dimension of a Euclidean space.
In this paper, we refer to tensor as geometric tensors,
and we consider a (discrete) rank-$p$ tensor field
$\tH\supp{p} \in \mathbb{R}^{\VV \times f \times d^p}$, where
$\VV$ denotes the number of vertices and
$f \in \mathbb{Z}^+$
($\Z^+$ denotes the positive integers).
Here, $f$ denotes the number of features (channels) of $\tH\supp{p}$, as shown in Figure~\ref{fig:tensors} (a).
With the indices, we denote $\etH\supp{p}_{i;g;k_1k_2 \dots k_p}$, where $i$ permutes
under the permutation of vertices and
$k_1, \dots, k_p$ refers to the Euclidean representation.
Thus, under the permutation,
$\pi: \etH\supp{p}_{i;g;k_1k_2 \dots k_p} \mapsto \etH\supp{p}_{\pi(i);g;k_1k_2 \dots k_p}$,
and under orthogonal transformation,
$\mU: \etH\supp{p}_{i;g;k_1k_2 \dots k_p} \mapsto \sum_{l_1, l_2, \dots, l_p} \emU_{k_1l_1}\emU_{k_2l_2}\dots\emU_{k_pl_p}\etH\supp{p}_{i;g;l_1l_2 \dots l_p}$.

\subsection{Construction of an isometric adjacency matrix}
\begin{wrapfigure}[15]{r}{0.5\textwidth}
  \centering
  \includegraphics[trim={0cm 43cm 35cm 0cm},clip,width=0.5\textwidth]
  {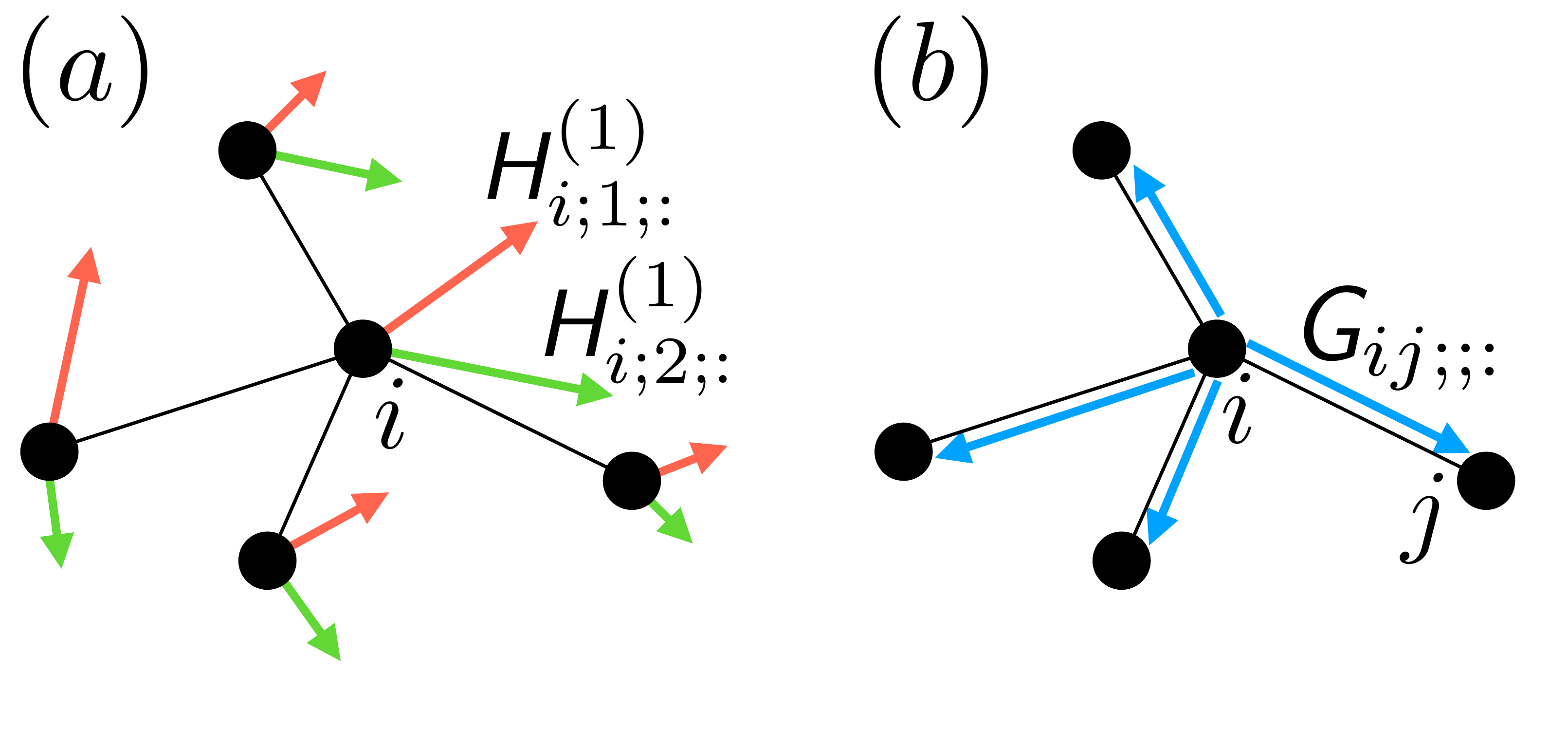}
  \caption{Schematic diagrams of
  (a) rank-1 tensor field $\tH\supp{1}$ with the number of features equaling 2 and
  (b) the simplest case of $\tG_{ij;;:} = \delta_{il}\delta_{jk}\emA_{ij} \mI (\vx_k - \vx_l) = \emA_{ij}(\vx_j - \vx_i)$.
  }
  \label{fig:tensors}
\end{wrapfigure}
Before constructing an IsoGCN, an \textbf{isometric adjacency matrix} (IsoAM),
which is at the core of the IsoGCN concept must be defined.
The proof of each proposition can be found in Appendix~\ref{sec:proofs}.

An IsoAM
$\tG \in \R^{\vert\V\vert \times \vert\V\vert \times d}$
is defined as:
\begin{align}
  \R^d \ni \tG_{ij;;:}
  := \sum_{k, l \in \V, k \neq l}\mT_{ijkl} (\vx_k - \vx_l),
  \label{eq:isoam}
\end{align}
where
$\tG_{ij;;:}$ is a slice in the spatial index of $\tG$,
$\vx_i \in \R^{d}$ is the position of the $i$th vertex (rank-1 tensor),
and $\mT_{ijkl} \in \R^{d \times d}$ is an untrainable transformation invariant and
orthogonal transformation equivariant rank-2 tensor.
Note that we denote $\etG_{ij;;k}$ to be consistent with
the notation of $\etH\supp{p}_{i;g;k_1k_2 \dots k_p}$
because $i$ and $j$ permutes under the vertex permutation
and $k$ represents the spatial index while the number of features is always 1.
The IsoAM can be viewed as a weighted adjacency matrix for each direction and
reflects spatial information while the usual weighted adjacency matrix
cannot because a graph has only one adjacency matrix.
If the size of the set $\{G_{ij;;:} \neq \bm{0}\}_j$ is greater than or equal to $d$, then
it can be deemed to be a frame, which is a generalization of a basis.
For the simplest case, one can define $\mT_{ijkl} = \delta_{il}\delta_{jk} \emA_{ij} \mI$ (Figure~\ref{fig:tensors} (b)),
where $\delta_{ij}$ is the Kronecker delta,
$\mA$ is the adjacency matrix of the graph,
and $\mI$ is the identity matrix that is the
simplest rank-2 tensor.
In another case, $\mT_{ijkl}$ can be determined from the geometry of a graph, as defined in~\eqref{eq:d}.
Nevertheless, in the bulk of this section, we retain $\mT_{ijkl}$ abstract to cover
various forms of interaction, such as position-aware GNNs~\citep{you2019position}.
Here, $\tG$ is composed of only untrainable parameters and thus can be determined before training.

\begin{proposition}\label{prop:gam_iso}
  IsoAM defined in~\eqref{eq:isoam} is
  both translation invariant and orthogonal transformation equivariant, i.e.,\
  for any isometric transformation
  $\forall \vt \in \R^3, \mU \in \O(d), T: \vx \mapsto \mU \vx + \vt$,
  \begin{align}
    T: \etG_{ij;;k} \mapsto \sum_{l} \emU_{kl} \etG_{ij;;l}.
  \end{align}
\end{proposition}

Based on the definition of the GCN layer in the equation~\ref{eq:gcn},
let $\tG \ast \tH\supp{0} \in \mathbb{R}^{\VV \times f \times d}$ denote the \textbf{convolution}
between $\tG$ and the rank-0 tensor field
$\tH\supp{0} \in \mathbb{R}^{\VV \times f}$ ($f \in \Z^+$)
as follows:
\begin{align}
  {(\tG \ast \tH\supp{0})}_{i;g;k} := \sum_j \tG_{ij;;k} \etH_{j;g;}\supp{0}\label{eq:convolution1}.
\end{align}

With a rank-1 tensor field
$\tH\supp{1} \in \R^{\VV \times f \times d}$,
let $\tG \odot \tH\supp{1} \in \R^{\VV \times f}$
and
$\tG \odot \tG \in \R^{\VV \times \VV}$
denote the \textbf{contractions}
which are defined as follows:
\begin{align}
  {(\tG \odot \tH\supp{1})}_{i;g;} := \sum_{j,k} \etG_{ij;;k} \etH\supp{1}_{j;g;k},
  \hspace{5pt}
  {(\tG \odot \tG)}_{il;;} := \sum_{j,k} \etG_{ij;;k} \etG_{jl;k}\label{eq:contraction_1}.
\end{align}
The contraction of IsoAMs
$\tG \odot \tG$ can be interpreted as the inner product of
each component in the IsoAMs.
Thus, the subsequent proposition follows.
\begin{proposition}\label{prop:gam_contraction}
  The contraction of IsoAMs $\tG \odot \tG$
  is isometric transformation invariant,
  i.e.,\ for any isometric transformation
  $\forall \vt \in \R^3, \mU \in \O(d), T: \vx \mapsto \mU \vx + \vt$,
  $\tG \odot \tG \mapsto \tG \odot \tG$.
\end{proposition}

With a rank-$p$ tensor field
$\tH\supp{p} \in \R^{\VV \times f \times d^p}$,
let
$\tG \otimes \tH\supp{p} \in \R^{\VV \times f \times d^{1+p}}$.
and
$\tG \otimes \tG \in \R^{\VV \times \VV \times d^2}$
denote the \textbf{tensor products}
defined as follows:
\begin{align}
  {(\tG \otimes \tH\supp{p})}_{i;g;km_1m_2 \dots m_p}
  &:= \sum_{j} \tG_{ij;;k} \etH\supp{p}_{j;g;m_1m_2 \dots m_p}\label{eq:tp_pq},
  \\
  {(\tG \otimes \tG)}_{il;;k_1k_2}
  &:= \sum_{j} \tG_{ij;;k_1} \tG_{jl;;k_2}.
\end{align}
The tensor product of IsoAMs $\tG \otimes \tG$
can be interpreted as the tensor product of each of the IsoAMs’ components.
Thus, the subsequent proposition follows:
\begin{proposition}\label{prop:gam_tensor}
  The tensor product of the IsoAMs $\tG \otimes \tG$
  is isometric transformation equivariant in terms of the rank-2 tensor,
  i.e.,\ for any isometric transformation
  $\forall \vt \in \R^3, \mU \in \O(d), T: \vx \mapsto \mU \vx + \vt$, and
  $\forall i, j \in {1, \dots, \VV}$,
  ${(\tG \otimes \tG)}_{ij;;k_1k_2} \mapsto \mU_{k_1l_1}\mU_{k_2l_2} {(\tG \otimes \tG)}_{ij;;l_1l_2}$.
\end{proposition}
This proposition is easily generalized to the tensors of higher ranks by
defining the $p$th tensor power of
$\tG$ as follows:
$\bigotimes^0 \tG = 1$,
$
\bigotimes^1 \tG = \tG
$,
and
$
\bigotimes^p \tG = \bigotimes^{p-1} \tG \otimes \tG
$.
Namely, $\bigotimes^p \tG$ is isometric transformation equivariant in terms of
rank-$p$ tensor.
Therefore, one can see that
$(\bigotimes^p \tG) \otimes \tH\supp{q} = (\bigotimes^{p-1} \tG) \otimes (\tG \otimes \tH\supp{q})$.
Moreover, the convolution can be generalized for $\bigotimes^p \tG$
and the rank-$0$ tensor field
$\tH\supp{0} \in \R^{\VV \times f}$
as follows:
\begin{align}
  {\left[ \left( \bigotimes^p \tG \right) \ast \tH\supp{0} \right]}_{i;g;k_1k_2 \dots k_p}
  =
  \sum_{j} \left(\bigotimes^p \tG \right)_{ij;;k_1k_2 \dots k_p} \emH\supp{0}_{j;g;}\label{eq:convolution_p}.
\end{align}
The contraction can be generalized for
$\bigotimes^p \tG$
and the rank-$q$ tensor field
$\tH\supp{q} \in \R^{\VV \times f \times d^q}$
($p \geq q$)
as specified below:
\begin{align}
  {\left[ \left( \bigotimes^p \tG \right) \odot \tH\supp{q} \right]}_{i;g;k_1k_2 \dots k_{p-q}}
  =
  \sum_{j, m_1, m_2, \dots, m_q}
  {\left( \bigotimes^p \tG \right)}_{ij;;k_1k_2 \dots k_{p-q}m_1m_2 \dots m_q} \etH\supp{q}_{j;g;m_1m_2 \dots m_q}\label{eq:contraction_p}.
\end{align}
For the case $p < q$, the contraction can be defined similarly.

\subsection{Construction of IsoGCN}
Using the operations defined above,
we can construct IsoGCN layers, which take
the tensor field of any rank as input, and output the tensor field of any rank,
which can differ from those of the input.
In addition, one can show that these layers are also equivariant under the vertex permutation,
as discussed in~\cite{maron2018invariant}.

\subsubsection{Isometric transformation invariant layer}
As can be seen in Proposition~\ref{prop:gam_iso}, the contraction of IsoAMs
is isometric transformation invariant.
Therefore, for the isometric transformation invariant layer with a rank-0 input tensor field
$f: \R^{\vert\V\vert \times f\sub{in}} \ni \mH\sub{in}\supp{0}
\mapsto \mH\sub{out}\supp{0} \in \R^{\vert\V\vert \times f\sub{out}}$
($f\sub{in}, f\sub{out} \in \Z^+$),
the activation function
$\sigma$,
and the trainable parameter matrix
$\mW \in \R^{f\sub{in} \times f\sub{out}}$
can be constructed as
$
\mH\sub{out}\supp{0} = \sigma\left(
\left( \tG \odot \tG \right) \mH\sub{in}\supp{0} \mW
\right)
$.
By defining
$\mL := \tG \odot \tG \in \R^{\vert\V\vert \times \vert\V\vert}$,
it can be simplified as
$
\mH\sub{out}\supp{0} = \sigma\left( \mL \mH\sub{in}\supp{0} \mW \right)
$,
which has the same form as a GCN (equation \ref{eq:gcn}),
with the exception that $\hat{\mA}$ is replaced with $\mL$.

An isometric transformation invariant layer with the rank-$p$ input tensor field
$\tH\supp{p}\sub{in} \in \mathbb{R}^{\VV \times f\sub{in} \times d^p}$
can be formulated as
$
\mH\sub{out}\supp{0} = F_{p \to 0}(\tH\sub{in}\supp{p}) = \sigma\left(
\left[ \bigotimes^p \tG \odot \tH\supp{p}\sub{in} \right] \mW \right).
$
If $p = 1$,
such approaches utilize the inner products of the vectors in $\R^d$,
these operations correspond to the extractions of a relative distance and an angle
of each pair of vertices,
which are employed in \citet{klicpera_dimenet_2020}.

\subsubsection{Isometric transformation equivariant layer}
To construct an isometric transformation equivariant layer,
one can use linear transformation, convolution and tensor product to the input tensors.
If both the input and the output tensor ranks are greater than 0,
one can apply neither nonlinear activation nor bias addition
because these operations will cause an inappropriate distortion of the isometry
because isometric transformation does not commute with them in general.
However, a conversion that uses only a linear transformation, convolution, and tensor product
does not have nonlinearity, which limits the predictive performance of the model.
To add nonlinearity to such a conversion, we can first convert the input tensors to
rank-0 ones, apply nonlinear activations, and then multiply them to the
higher rank tensors.

The nonlinear isometric transformation equivariant layer with
the rank-$m$ input tensor field
$\tH\sub{in}\supp{m} \in \mathbb{R}^{\VV \times f\sub{in} \times d^m}$
and the rank-$l$ ($m \leq l$) output tensor
$\tH\sub{out}\supp{l} \in \mathbb{R}^{\VV \times f\sub{out} \times d^l}$
can be defined as:
\begin{align}
  \tH\sub{out}\supp{l}
  = F_{m \to 0}\left(\tH\sub{in}\supp{m}\right) \times \tF_{m \to l}\left(\tH\sub{in}\supp{m}\right),
  \hspace{10pt}
  \tF_{m \to l}\left(\tH\sub{in}\supp{m}\right) =
  \left[ \bigotimes^{l - m} \tG \right] \otimes \tH\sub{in}\supp{m} \mW^{ml},
  \label{eq:isogcn_mleql}
\end{align}
where $\times$ denotes multiplication with broadcasting
and $\mW^{ml} \in \R^{f\sub{in} \times f\sub{out}}$ are trainable weight matrices
multiplied in the feature direction.
If $m = 0$, we regard $\tG \otimes \tH\supp{0}$ as $\tG \ast \tH\supp{0}$.
If $m = l$, one can add the residual connection~\citep{he2016deep} in equation~\ref{eq:isogcn_mleql}.
If $m > l$,
\begin{align}
  \tH\sub{out}\supp{l}
  = F_{m \to 0}\left(\tH\sub{in}\supp{m}\right) \times \tF_{m \to l}\left(\tH\sub{in}\supp{m}\right),
  \hspace{10pt}
  \tF_{m \to l}\left(\tH\sub{in}\supp{m}\right) =
  \left[ \bigotimes^{m - l} \tG \right] \odot \tH\sub{in}\supp{m} \mW^{ml}.
\end{align}
In general, the nonlinear isometric transformation equivariant layer with the rank-0 to
rank-$M$ input tensor field $\{\tH\sub{in}\supp{m}\}_{m=0}^M$
and the rank-$l$ output tensor field $\tH\sub{out}\supp{l}$ can be defined as:
\begin{align}
  \tH\supp{l}_{\mathrm{out}}
  &= \tH\supp{l}_{\mathrm{in}} \mW
  + \sum_{m=0}^M f\sub{gather}\left( \left\{ F_{k\to0}(\tH\sub{in}\supp{k}) \right\}_{k=0}^M \right)
  \times \tF_{m \to l} \left( \tH\sub{in}\supp{m} \right),
  \label{eq:isogcn}
\end{align}
where $f_\mathrm{gather}$ denotes a function
such as summation, product and concatenation in the feature direction.
One can see that this layer is similar to that in the TFN (equation~\ref{eq:tfn}),
while there are no spherical harmonics and trainable message passing.

To be exact, the output of the layer defined above is translation invariant.
To output translation equivariant variables such as the vertex positions after deformation
(which change accordingly with the translation of the input graph),
one can first define the reference vertex position $\vx\sub{ref}$ for each graph,
then compute the translation invariant output using equation~\ref{eq:isogcn},
and finally, add $\vx\sub{ref}$ to the output.
For more detailed information on IsoGCN modeling, see Appendix~\ref{sec:modeling}.

\subsection{Example of IsoAM}\label{sec:concrete_isoam}
The IsoGCN $\tG$ is defined in a general form
for the propositions to work with various classes of graph.
In this section, we concretize the concept of the IsoAM
to apply an IsoGCN to mesh structured data.
Here, a mesh is regarded as a graph regarding the points in the mesh as vertices
of the graph and assuming two vertices are connected when they share the same cell.
A concrete instance of IsoAM
$\tilde{\tD}, \tD \in \R^{\vert\V\vert \times \vert\V\vert \times d}$
is defined as follows:
\begin{wraptable}[8]{r}[0pt]{0.4\textwidth}
  \centering
  \caption{Correspondence between the differential operators and
  the expressions using the IsoAM $\tilde{\tD}$.}
  \begin{tabular*}{0.4\textwidth}{ll}
    \textbf{Differential op.} & \textbf{Expression}
    \\
    \hline
    Gradient & $\tilde{\tD} \ast \tH\supp{0}$
    \\
    Divergence & $\tilde{\tD} \odot \tH\supp{1}$
    \\
    Laplacian & $\tilde{\tD} \odot \tilde{\tD} \tH\supp{0}$
    \\
    Jacobian & $\tilde{\tD} \otimes \tH\supp{1}$
    \\
    Hessian & $\tilde{\tD} \otimes \tilde{\tD} \ast \tH\supp{0}$
  \end{tabular*}
  \label{tab:differential}
\end{wraptable}
\begin{align}
  \tilde{\etD}_{ij;;k} &= \etD_{ij;;k} - \delta_{ij} \sum_l \etD_{il;;k},
  \label{eq:tilde_d}
  \\
  \etD_{ij;;:} &=
  \mM^{-1}_{i} \frac{\vx_{j} - \vx_{i}}{\lVert\vx_{j} - \vx_{i}\rVert^2} w_{ij} \emA_{ij}(m),
  \label{eq:d}
  \\
  \mM_i &= \sum_l \frac{\vx_l - \vx_i}{\lVert \vx_{l} - \vx_{i} \rVert}
  \otimes \frac{\vx_{l} - \vx_{i}}{\lVert \vx_{l} - \vx_{i} \rVert} w_{il} \emA_{il}(m),
\end{align}
where
$\R^{\vert\V\vert \times \vert\V\vert} \ni \mA(m) := \min{(\sum_{k=1}^{m} \mA^k, 1)}$
is an adjacency matrix up to $m$ hops
and $w_{ij} \in \R$ is an untrainable weight between the $i$th and $j$th vertices
that is determined depending on the tasks\footnote{$\mM_i$ is invertible when the number of independent vectors in
$\{\vx_{l} - \vx_{i}\}_l$
is greater than or equal to the space dimension $d$, which is true
for common meshes, e.g., a solid mesh in 3D Euclidean space.}.
By regarding
$\mT_{ijkl} = \delta_{il} \delta_{jk} \mM_i^{-1} w_{ij} \mA_{ij}(m) / \Vert \vx_j - \vx_i \Vert^2$
in equation~\ref{eq:isoam},
one can see that $\tD$ is qualified as an IsoAM.
Because a linear combination of IsoAMs is also an IsoAM, $\tilde{\tD}$ is an IsoAM.
Thus, they provide both translation invariance and orthogonal transformation equivariance.
$\tilde{\tD}$ can be obtained only from the mesh geometry information,
thus can be computed in the preprocessing step.

Here, $\tilde{\tD}$ is designed such that it corresponds to
the gradient operator model used in physical simulations \citep{tamai2014least, swartz1969generalized}.
As presented in Table~\ref{tab:differential} and Appendix~\ref{sec:operators},
$\tilde{\tD}$ is closely related to many differential operators,
such as the gradient, divergence, Laplacian, Jacobian, and Hessian.
Therefore, the considered IsoAM plays an essential role in constructing
neural network models that are capable of learning differential equations.

\section{Experiment}
To test the applicability of the proposed model,
we composed the following two datasets:
1) a differential operator dataset of grid meshes; and
2) an anisotropic nonlinear heat equation dataset of meshes generated from CAD data.
In this section, we discuss our machine learning model, the definition of the problem, and the results for each dataset.

Using $\tilde{\tD}$ defined in Section~\ref{sec:concrete_isoam},
we constructed a neural network model considering
an encode-process-decode configuration~\citep{battaglia2018relational}.
The encoder and decoder were comprised of component-wise MLPs and tensor operations.
For each task, we tested $m = 2, 5$ in \eqref{eq:d}
to investigate the effect of the number of hops considered.
In addition to the GCN~\citep{kipf2017semi},
we chose
GIN~\citep{xu2018powerful},
SGCN~\citep{pmlr-v97-wu19e},
Cluster-GCN~\citep{chiang2019cluster}, and
GCNII~\citep{chen2020simple}
as GCN variant baseline models.
For the equivariant models, we chose
the TFN~\citep{thomas2018tensor} and
SE(3)-Transformer~\citep{fuchs2020se} as the baseline.
We implemented these models using PyTorch 1.6.0~\citep{NEURIPS2019_9015}
and PyTorch Geometric 1.6.1~\citep{Fey/Lenssen/2019}.
For both the TFN and SE(3)-Transformer, we used implementation of \citet{fuchs2020se}
\footnote{\url{https://github.com/FabianFuchsML/se3-transformer-public}}
because the computation of the TFN is considerably faster than the original implementation,
as claimed in \citet{fuchs2020se}.
For each experiment, we minimized the mean squared loss using the Adam optimizer~\citep{kingma2014adam}.
The corresponding implementation and the dataset will be made available online.
The details of the experiments can be found in Appendix~\ref{sec:differential} and~\ref{sec:nl_tensor}.

\begin{figure}[tb]
  \centering
  \stackunder[-1pt]
  {\includegraphics[trim={30cm 5cm 30cm 5cm},clip,width=0.18\textwidth]
  {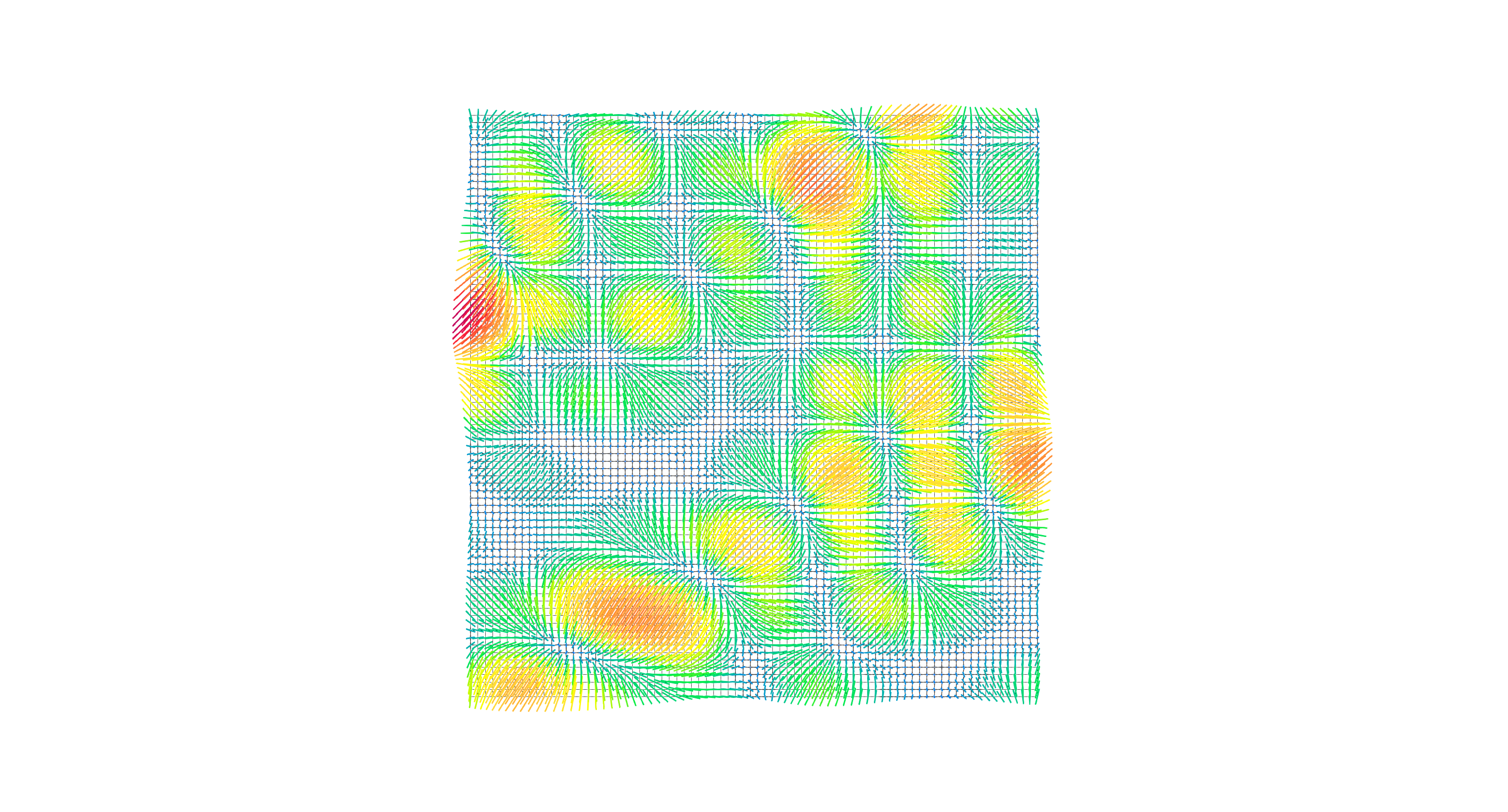}
  }{Ground truth}
  \stackunder[-1pt]
  {\includegraphics[trim={30cm 5cm 30cm 5cm},clip,width=0.18\textwidth]
  {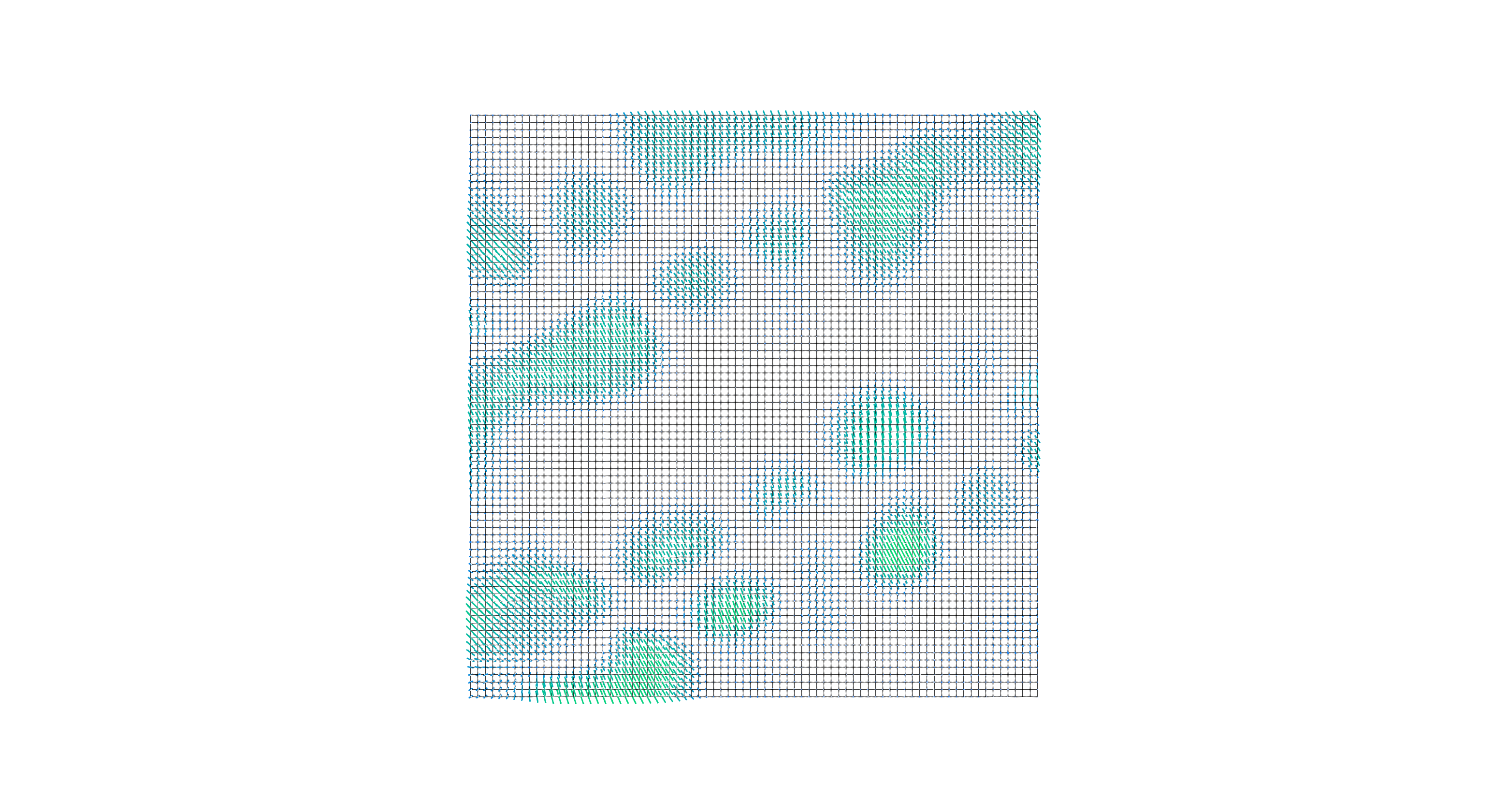}
  }{}
  \stackunder[-1pt]
  {\includegraphics[trim={30cm 5cm 30cm 5cm},clip,width=0.18\textwidth]
  {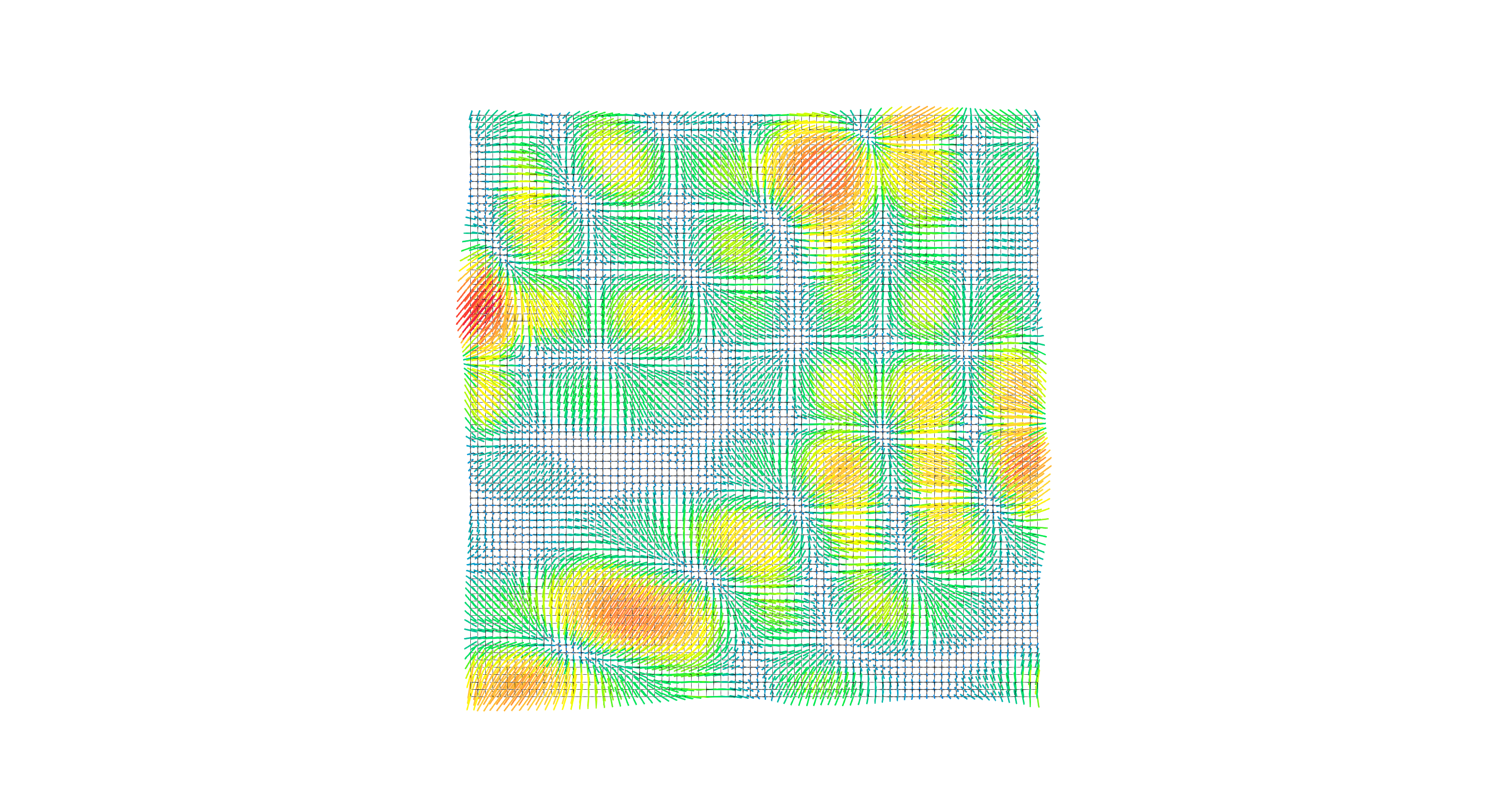}
  }{}
  \stackunder[-1pt]
  {\includegraphics[trim={30cm 5cm 30cm 5cm},clip,width=0.18\textwidth]
  {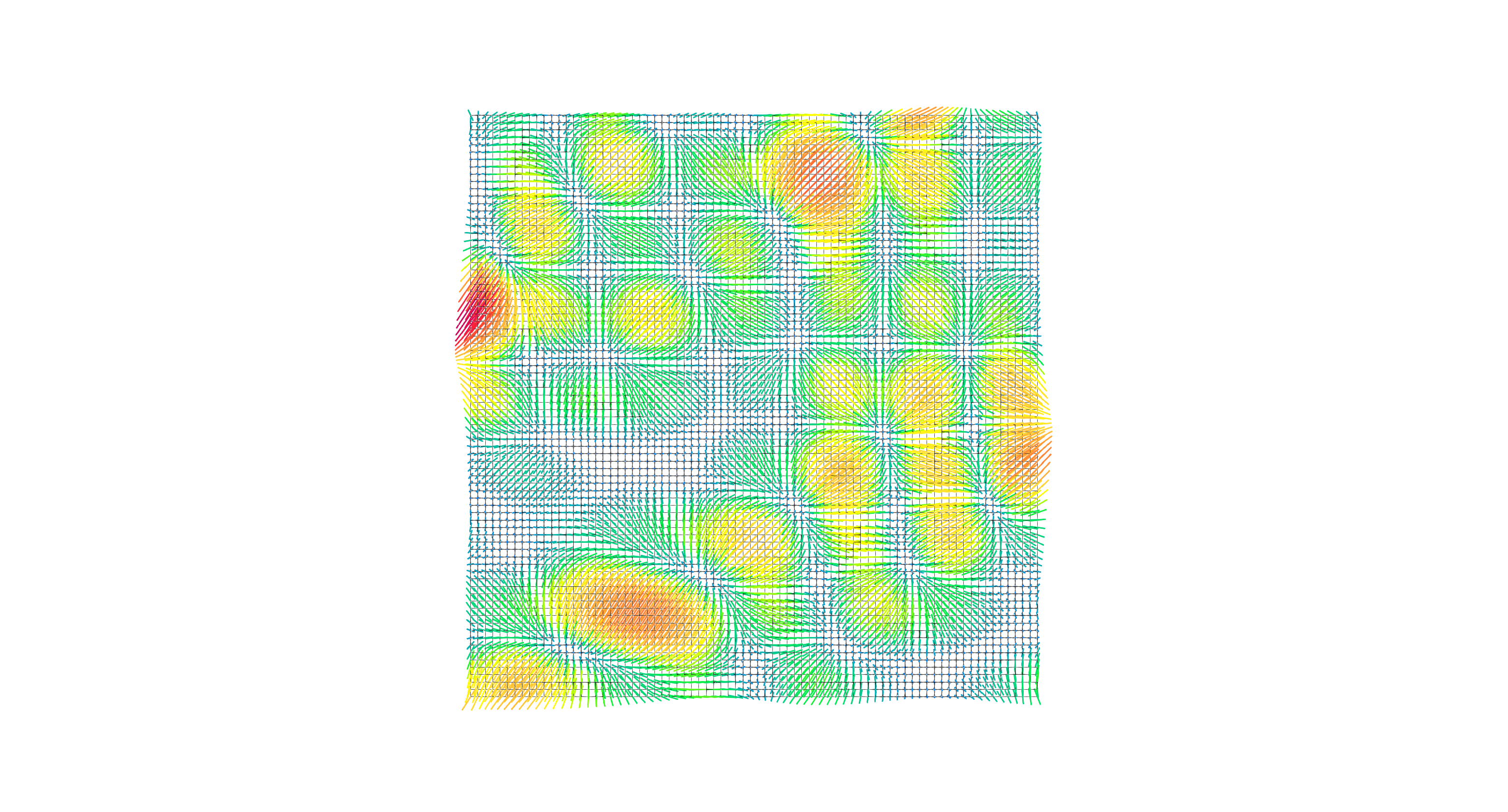}
  }{}
  {\includegraphics[width=0.1\textwidth]{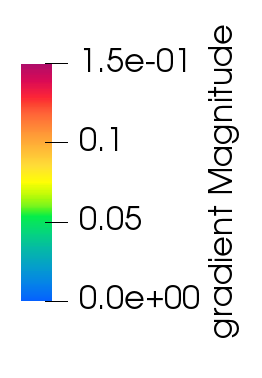}}
  \\[-20pt]
  \stackunder[-1pt]
  {\includegraphics[trim={30cm -10cm 30cm 60cm},clip,width=0.18\textwidth]
  {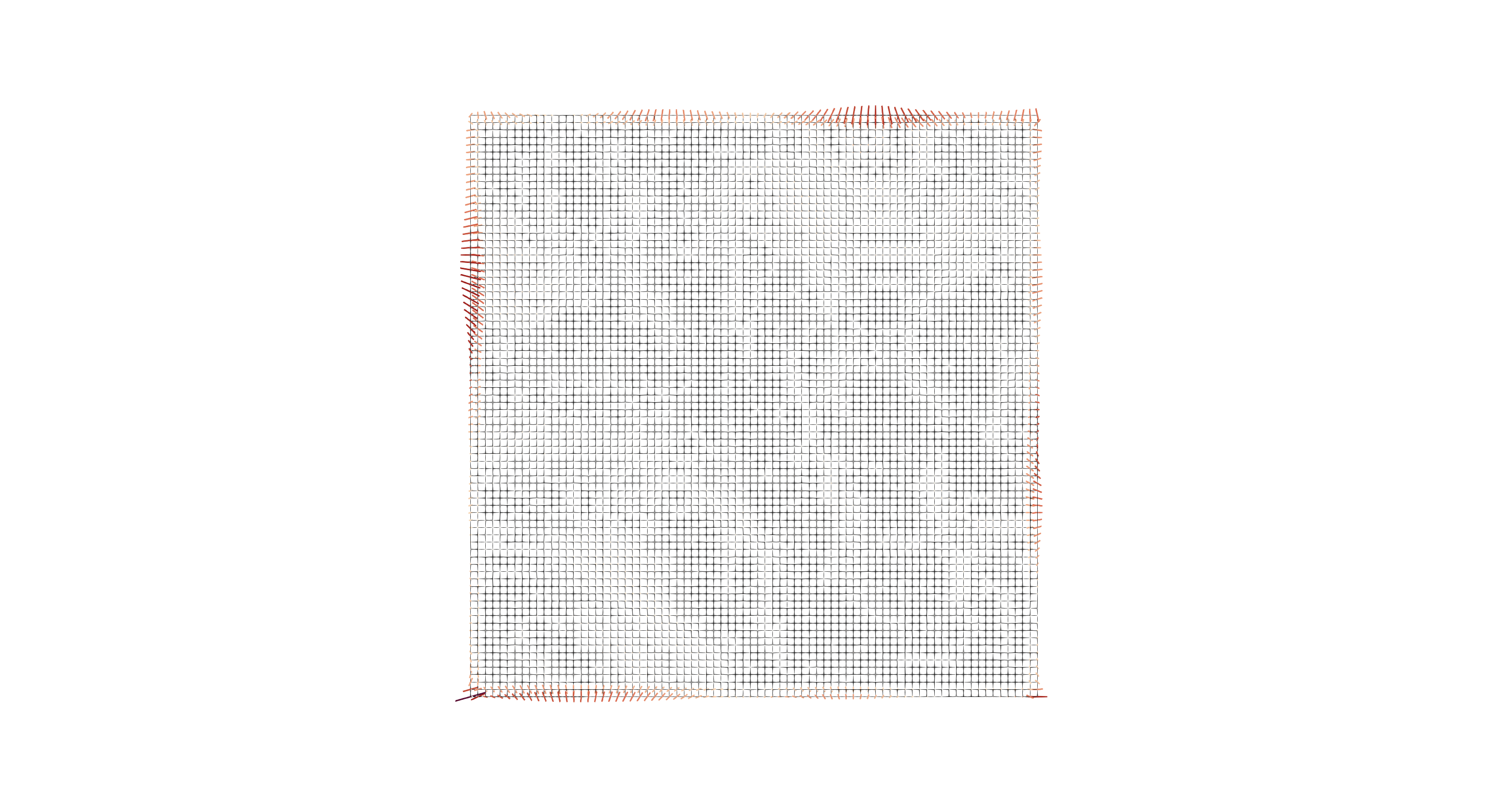}
  }{}
  \stackunder[-1pt]
  {\includegraphics[trim={30cm 5cm 30cm 5cm},clip,width=0.18\textwidth]
  {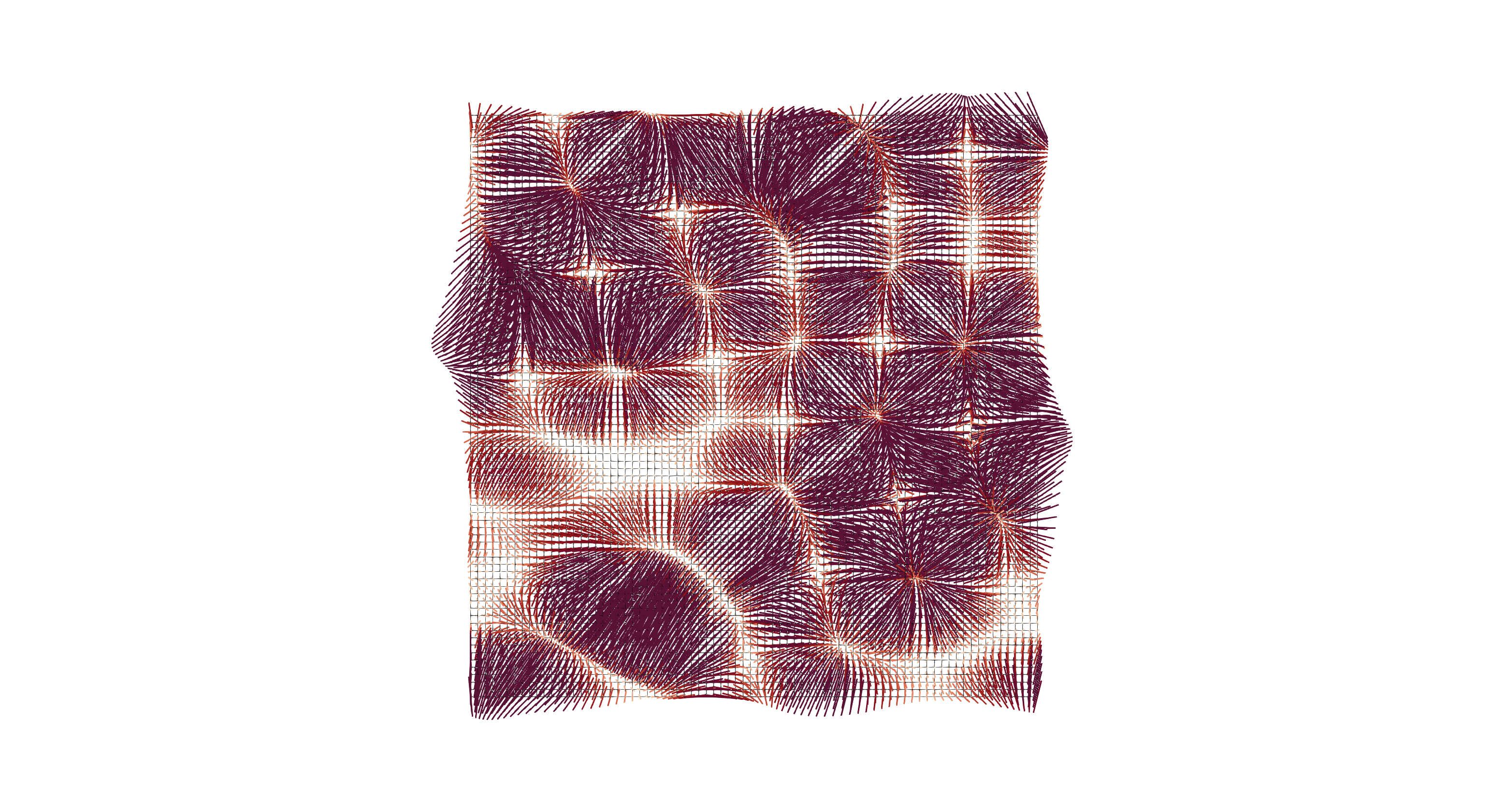}
  }{Cluster-GCN}
  \stackunder[-1pt]
  {\includegraphics[trim={30cm 5cm 30cm 5cm},clip,width=0.18\textwidth]
  {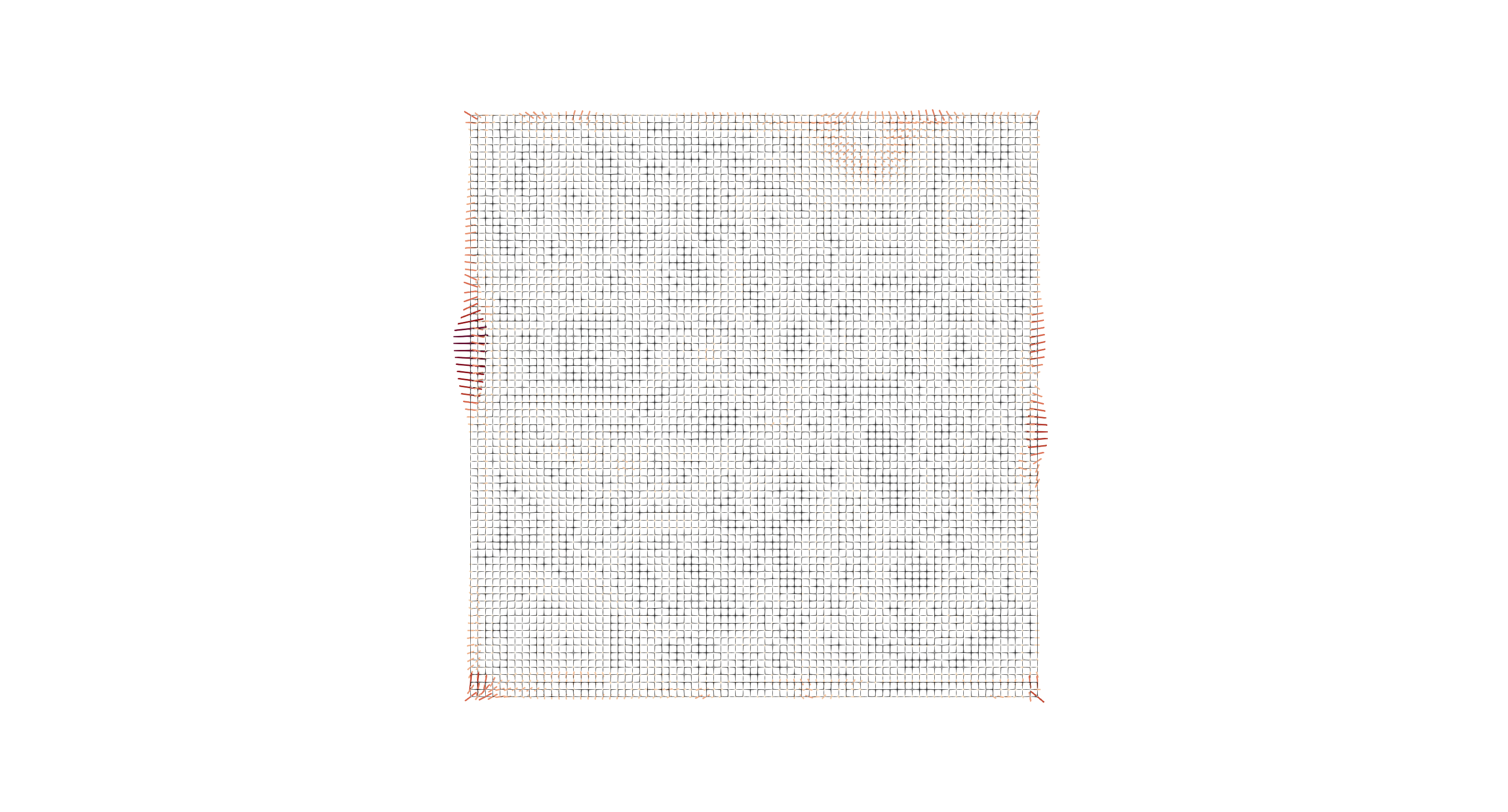}
  }{SE(3)-Transformer}
  \stackunder[-1pt]
  {\includegraphics[trim={30cm 5cm 30cm 5cm},clip,width=0.18\textwidth]
  {figs/grid/iso_gcn_adj2_factor_2020-09-30_08-16-08.511597/grid_difference.png}
  }{IsoGCN (Ours)}
  {\includegraphics[width=0.1\textwidth]{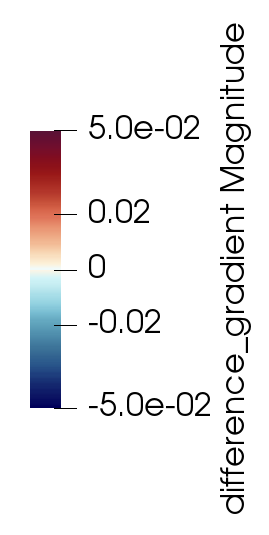}}
  \caption{
    (Top) the gradient field and (bottom) the error vector between the prediction and the ground truth
    of a test data sample.
    The error vectors are exaggerated by a factor of 2 for clear visualization.
  }
  \label{fig:gradient}
\end{figure}
\bgroup
\def\arraystretch{0.9}
\begin{table}[bt]
  \centering
  \caption{Summary of the test losses (mean squared error
  $\pm$ the standard error of the mean in the original scale)
  of the differential operator dataset:
  $0 \rightarrow 1$
  (the scalar field to the gradient field),
  $0 \rightarrow 2$
  (the scalar field to the Hessian field),
  $1 \rightarrow 0$
  (the gradient field to the Laplacian field),
  and $1 \rightarrow 2$
  (the gradient field to the Hessian field).
  Here, if ``$\vx$'' is ``Yes'', $\vx$ is also in the input feature.
  We show only the best setting for each method except for the equivariant models.
  For a full table, see Appendix~\ref{sec:differential}.
  OOM denotes the out-of-memory on the applied GPU (32 GiB).}
  \scalebox{0.9}{
  \begin{tabular}{lcccccc}
    \textbf{Method} & \textbf{\# hops} & \textbf{$\vx$}
    & \makecell{\textbf{Loss of }\boldsymbol{$0 \to 1$}\\$\times 10^{-5}$}
    & \makecell{\textbf{Loss of }\boldsymbol{$0 \to 2$}\\$\times 10^{-6}$}
    & \makecell{\textbf{Loss of }\boldsymbol{$1 \to 0$}\\$\times 10^{-6}$}
    & \makecell{\textbf{Loss of }\boldsymbol{$1 \to 2$}\\$\times 10^{-6}$}
    \\
    \hhline{=======}
    GIN & 5 & Yes & 147.07 $\pm$ 0.51 & 47.35 $\pm$ 0.35 & 404.92 $\pm$ 1.74 & 46.18 $\pm$ 0.39
    \\
    GCNII & 5 & Yes & 151.13 $\pm$ 0.53 & 31.87 $\pm$ 0.22 & 280.61 $\pm$ 1.30 & 39.38 $\pm$ 0.34
    \\
    SGCN & 5 & Yes & 151.16 $\pm$ 0.53 & 55.08 $\pm$ 0.42 & 127.21 $\pm$ 0.63 & 56.97 $\pm$ 0.44
    \\
    GCN & 5 & Yes & 151.14 $\pm$ 0.53 & 48.50 $\pm$ 0.35 & 542.30 $\pm$ 2.14 & 25.37 $\pm$ 0.28
    \\
    Cluster-GCN & 5 & Yes & 146.91 $\pm$ 0.51 & 26.60 $\pm$ 0.19 & 185.21 $\pm$ 0.99 & 18.18 $\pm$ 0.20
    \\
    \hline
    \multirow{2}{*}{TFN}
    & 2 & No & 2.47 $\pm$ 0.02 & OOM & 26.69 $\pm$ 0.24 & OOM
    \\
    & 5 & No & OOM & OOM & OOM & OOM
    \\[3pt]
    \multirow{2}{*}{SE(3)-Trans.}
    & 2 & No & \textbf{1.79} $\pm$ 0.02 & \textbf{3.50} $\pm$ 0.04 & \textbf{2.52} $\pm$ 0.02 & OOM
    \\
    & 5 & No & 2.12 $\pm$ 0.02 & OOM & 7.66 $\pm$ 0.05 & OOM
    \\[3pt]
    \multirow{2}{*}{\textbf{IsoGCN} (Ours)}
    & 2 & No & 2.67 $\pm$ 0.02 & 6.37 $\pm$ 0.07 & 7.18 $\pm$ 0.06 & \textbf{1.44} $\pm$ 0.02
    \\
    & 5 & No & 14.19 $\pm$ 0.10 & 21.72 $\pm$ 0.25 & 34.09 $\pm$ 0.19 & 8.32 $\pm$ 0.09
  \end{tabular}
  }
  \label{tab:differential_dataset_results}
\end{table}
\egroup

\subsection{Differential operator dataset}
To demonstrate the expressive power of IsoGCNs, we created a dataset to learn
the differential operators.
We first generated a pseudo-2D grid mesh randomly with only one cell in the $Z$ direction
and 10 to 100 cells in the $X$ and $Y$ directions.
We then generated scalar fields on the grid meshes
and analytically calculated the gradient, Laplacian, and Hessian fields.
We generated 100 samples for each train, validation, and test dataset.
For simplicity, we set $w_{ij} = 1$ in \eqref{eq:d} for all $(i, j) \in \E$.
To compare the performance with the GCN models, we simply replaced an IsoGCN layer with
a GCN or its variant layers while keeping the
number of hops $m$ the same to enable a fair comparison.
We adjusted the hyperparameters for the equivariant models to ensure that
the number of parameters in each was almost the same as that in the IsoGCN model.
For more details regarding the model architecture, see Appendix~\ref{sec:differential}.
We conducted the experiments using the following settings:
1) inputting the scalar field and predicting the gradient field (rank-0 $\rightarrow$ rank-1 tensor);
2) inputting the scalar field and predicting the Hessian field (rank-0 $\rightarrow$ rank-2 tensor);
3) inputting the gradient field and predicting the Laplacian field (rank-1 $\rightarrow$ rank-0 tensor); and
4) inputting the gradient field and predicting the Hessian field (rank-1 $\rightarrow$ rank-2 tensor).

Figure~\ref{fig:gradient} and Table~\ref{tab:differential_dataset_results}
present a visualization and comparison of predictive performance, respectively.
The results show that an IsoGCN outperforms other GCN models for all settings.
This is because the IsoGCN model has information on the relative position of the adjacency vertices,
and thus understands the direction of the gradient, whereas the
other GCN models cannot distinguish where the adjacencies are, making it nearly impossible to
predict the gradient directions.
Adding the vertex positions to the input feature to other GCN models
exhibited a performance improvement, however as the vertex position is
not a translation invariant feature, it could degrade the predictive
performance of the models.
Thus, we did not input $\vx$ as a vertex feature to the IsoGCN model or other equivariant models
to retain their isometric transformation invariant and equivariant natures.
IsoGCNs perform competitively against other equivariant models
with shorter inference time as shown in Table~\ref{tab:differential_speed}.
As mentioned in Section~\ref{sec:concrete_isoam}, $\tilde{\tD}$ corresponds to
the gradient operator, which is now confirmed in practice.

\subsection{Anisotropic nonlinear heat equation dataset}
To apply the proposed model to a real problem, we adopted the
anisotropic nonlinear heat equation.
We considered the task of predicting the time evolution of the temperature field
based on the initial temperature field, material property, and mesh geometry information as inputs.
We randomly selected 82 CAD shapes from the first 200 shapes of the ABC
dataset~\citep{Koch_2019_CVPR}, generate first-order tetrahedral meshes using
a mesh generator program, Gmsh~\citep{gauzaine2009},
randomly set the initial temperature and anisotropic thermal conductivity,
and finally conducted a finite element analysis (FEA) using the FEA program
FrontISTR\footnote{\url{https://github.com/FrontISTR/FrontISTR}. We applied a private update to FrontISTR to deal with the anisotropic heat problem, which will be also made available online.}~\citep{morita2016development, ihara2017web}.

For this task, we set $w_{ij} = V_j\sup{effective} / V_i\sup{effective}$, where
$\displaystyle V_i\sup{effective}$
denotes the effective volume of the $i$th vertex (\eqref{eq:v_effective}.)
Similarly to the differential operator dataset, we tested the number of hops $m = 2, 5$.
However because we put four IsoAM operations in one model,
the number of hops visible from the model is 8 ($m = 2$) or 20 ($m = 5$).
As is the case with the differential operator dataset, we replaced an IsoGCN layer
accordingly for GCN or its variant models.
In the case of $k = 2$, we reduced the number of parameters for each of the equivariant models
to fewer than the IsoGCN model because they exceeded the memory of the GPU (NVIDIA Tesla V100 with 32 GiB memory) with the same number of parameters.
In the case of $k = 5$, neither the TFN nor the SE(3)-Transformer fits into the memory of the GPU
even with the number of parameters equal to 10.
For more details about the dataset and the model, see Appendix~\ref{sec:nl_tensor}.

\begin{figure}[tb]
  \centering
  \stackunder[-1pt]
  {\includegraphics[trim={32cm 3cm 36cm 3cm},clip,width=0.15\textwidth]
  {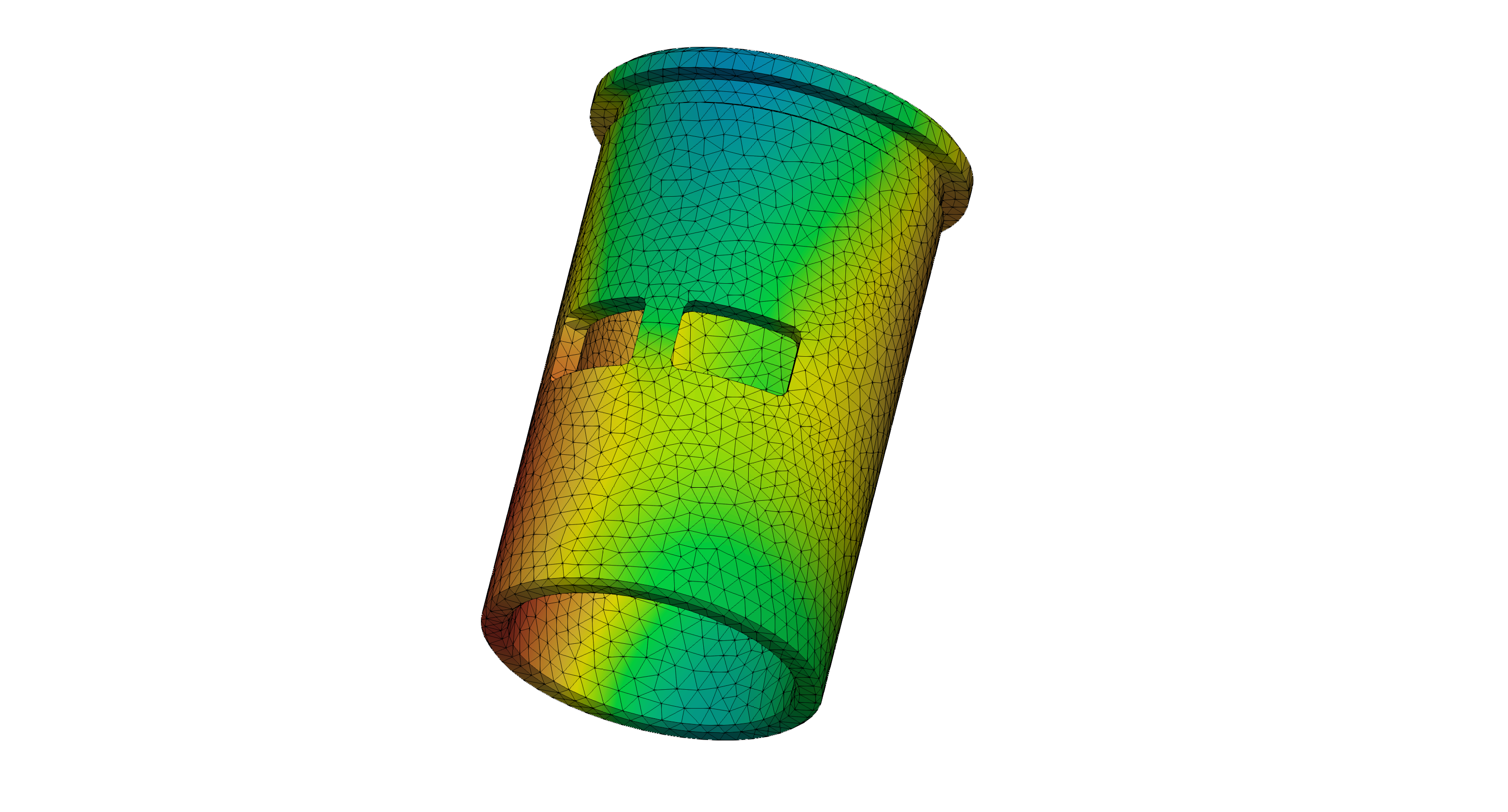}
  }{Ground truth}
  \stackunder[-1pt]
  {\includegraphics[trim={32cm 3cm 36cm 3cm},clip,width=0.15\textwidth]
  {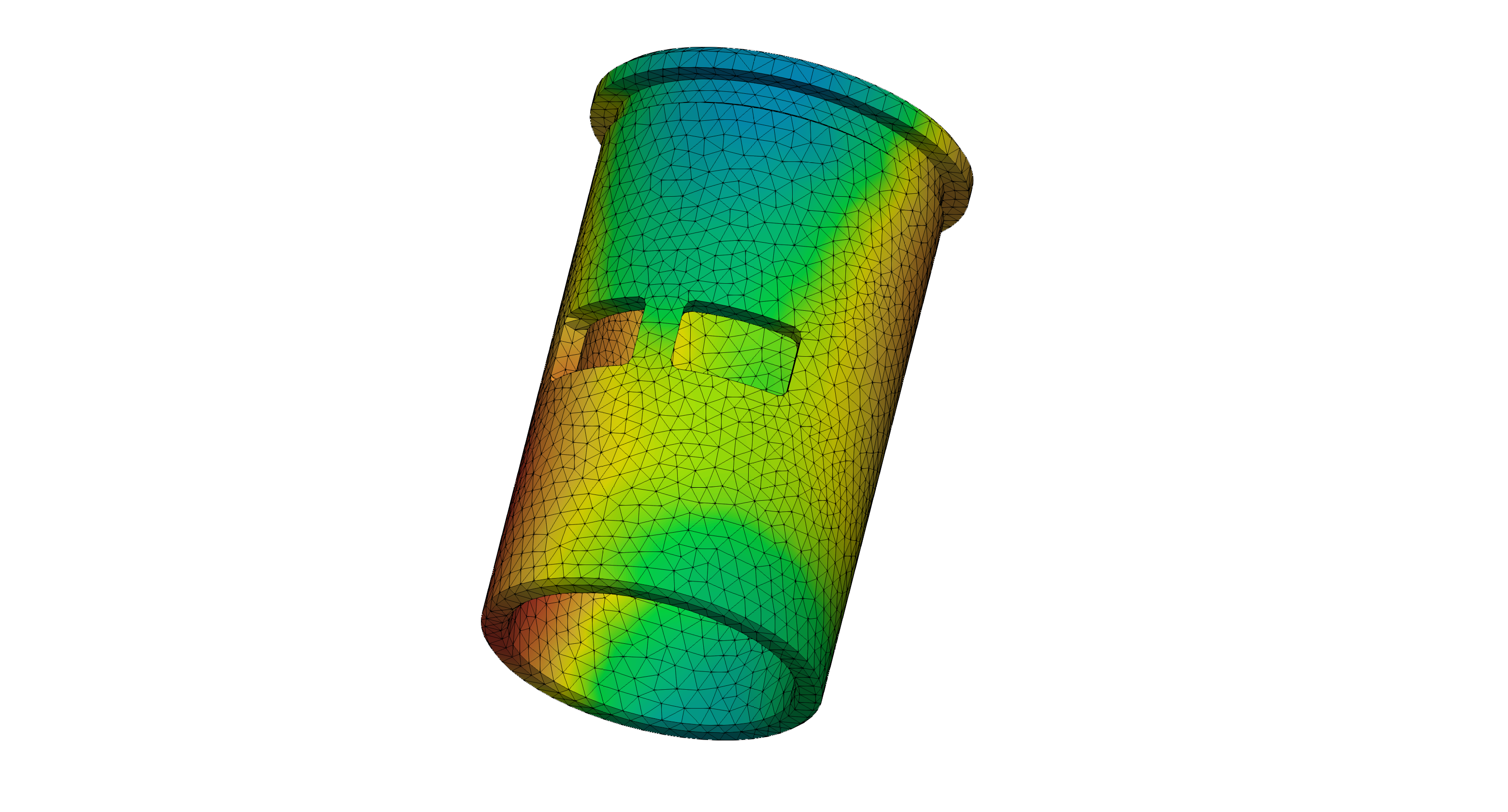}
  }{}
  \stackunder[-1pt]
  {\includegraphics[trim={32cm 3cm 36cm 3cm},clip,width=0.15\textwidth]
  {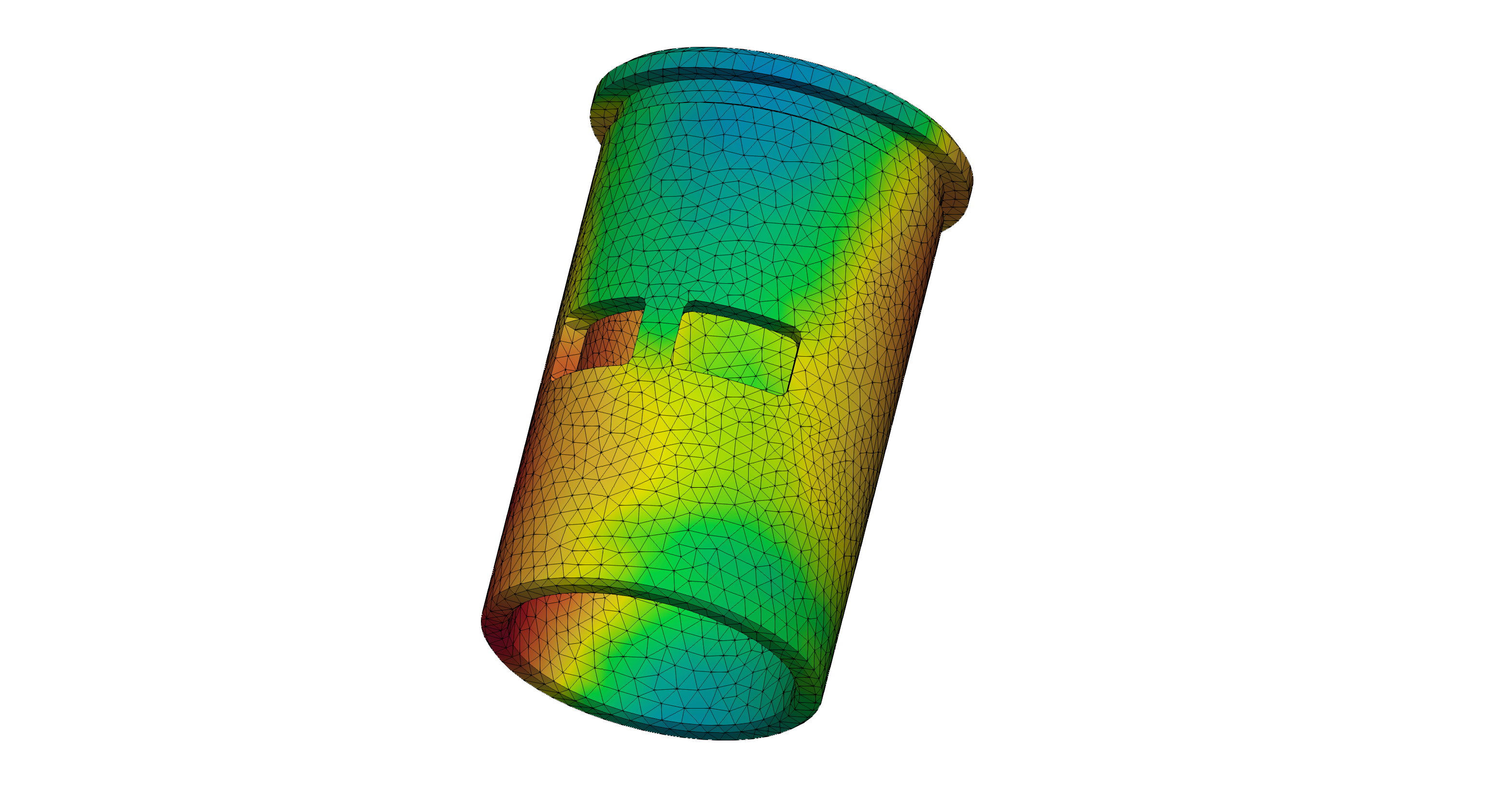}
  }{}
  \stackunder[-1pt]
  {\includegraphics[trim={32cm 3cm 36cm 3cm},clip,width=0.15\textwidth]
  {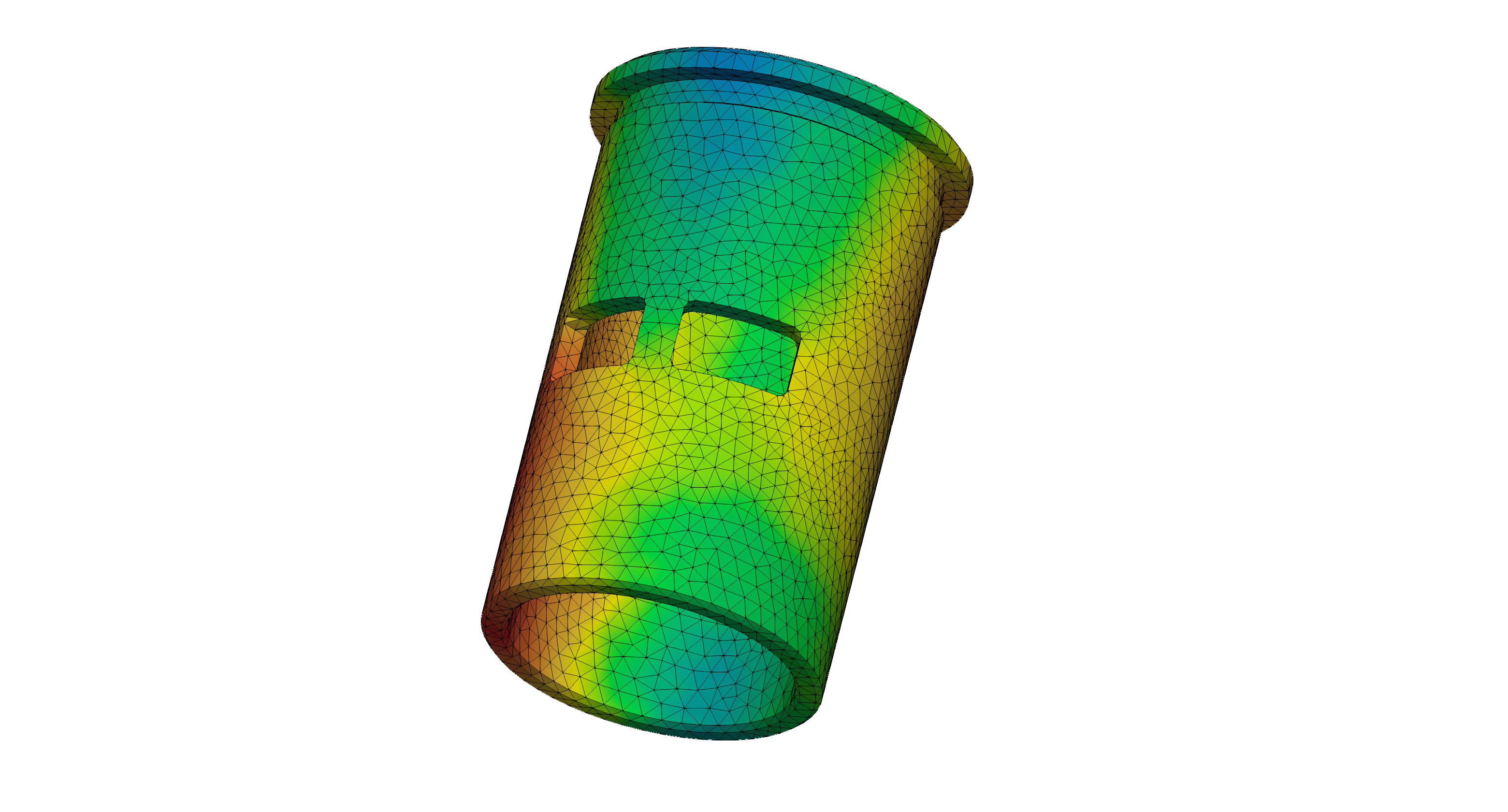}
  }{}
  {\includegraphics[trim={0cm -1cm 0cm 0cm},clip,width=0.09\textwidth]{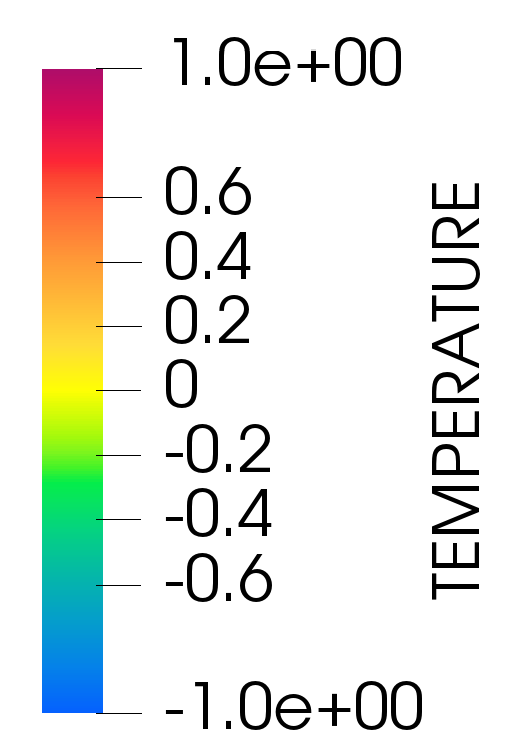}}
  \\
  \stackunder[-1pt]
  {\includegraphics[trim={32cm -20cm 36cm 80cm},clip,width=0.15\textwidth]
  {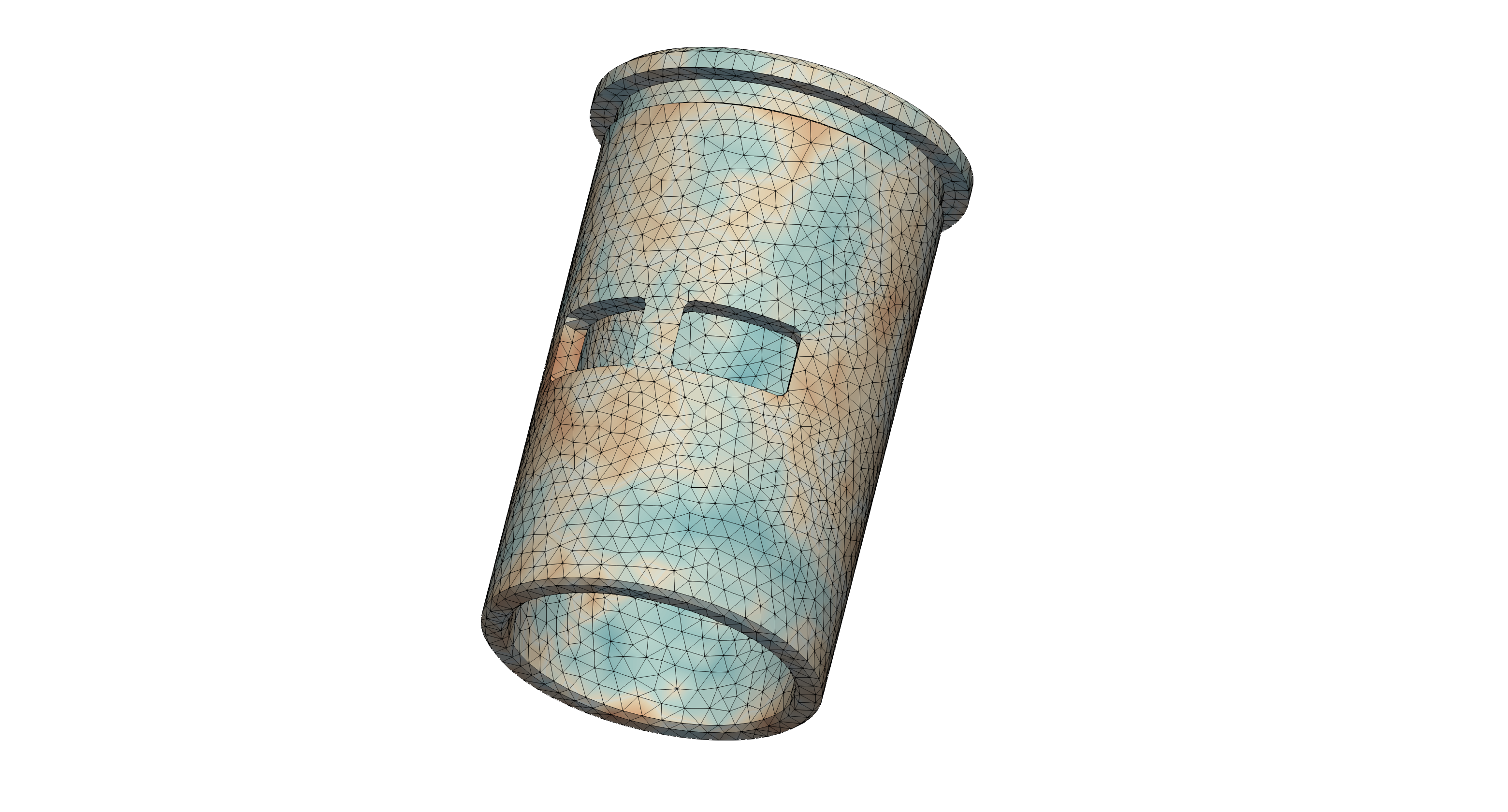}
  }{}
  \stackunder[-1pt]
  {\includegraphics[trim={32cm 3cm 36cm 3cm},clip,width=0.15\textwidth]
  {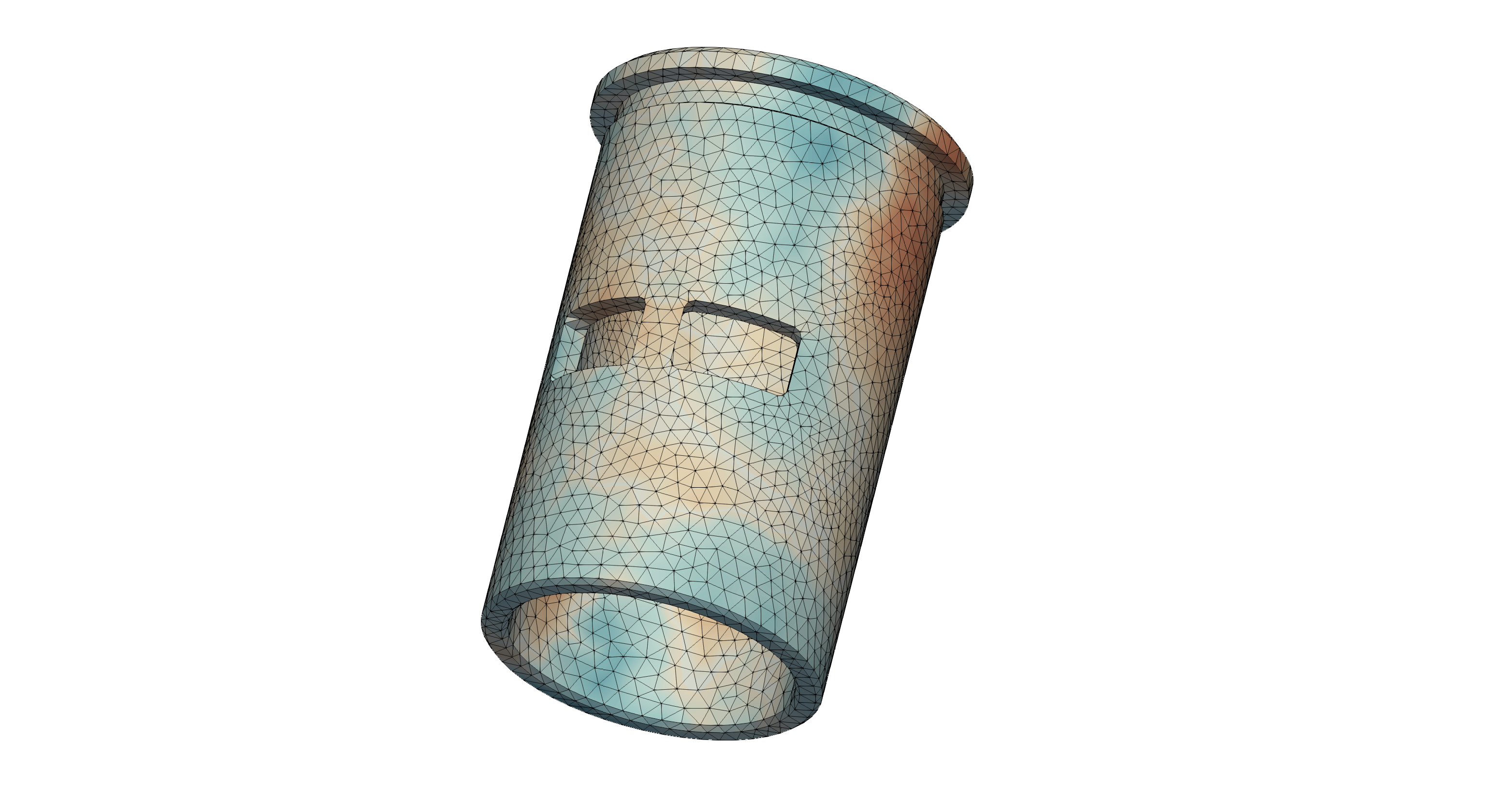}
  }{SGCN\hspace{25pt}}
  \stackunder[-1pt]
  {\includegraphics[trim={32cm 3cm 36cm 3cm},clip,width=0.15\textwidth]
  {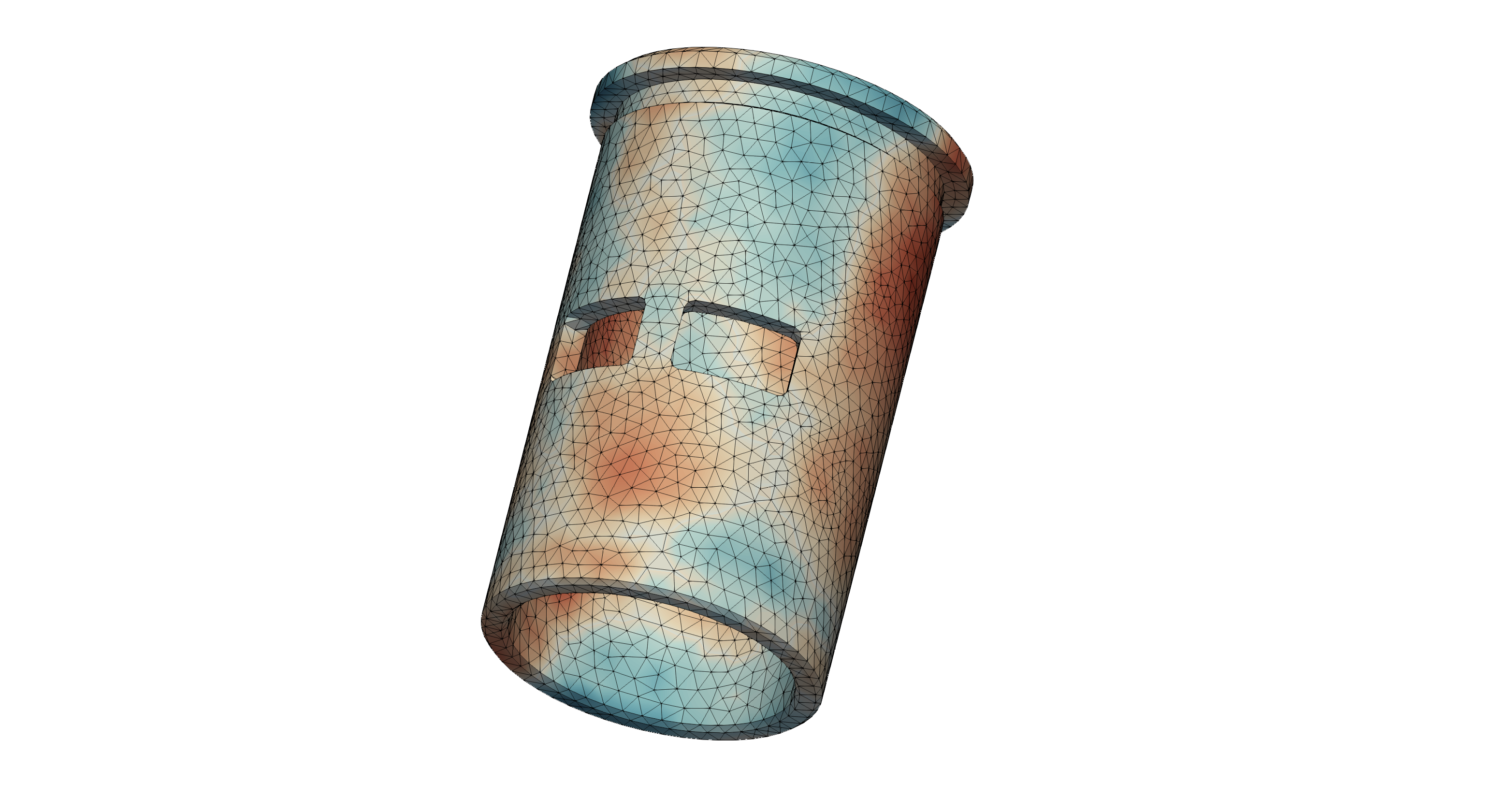}
  }{SE(3) Trans.}
  \stackunder[-1pt]
  {\includegraphics[trim={32cm 3cm 36cm 3cm},clip,width=0.15\textwidth]
  {figs/nl_tensor/00000111/clscale0.25/steepness1.0_rep0/isogcn_adj2_2020-09-30_09-31-56.747461/nl_tensor_difference.png}
  }{IsoGCN (Ours)}
  {\includegraphics[trim={0cm -1cm 0cm 0cm},clip,width=0.1\textwidth]{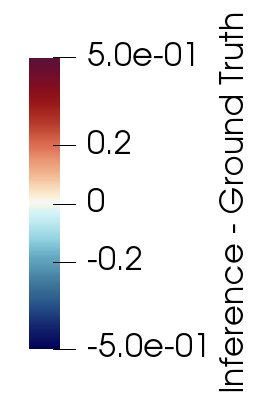}}
  \caption{(Top) the temperature field of the ground truth and inference results
  and (bottom) the error between the prediction and the ground truth
  of a test data sample.
  The error is exaggerated by a factor of 2 for clear visualization.}
  \label{fig:nl_tensor}
\end{figure}

Figure~\ref{fig:nl_tensor} and Table~\ref{tab:nl_tensor} present the results of the qualitative and quantitative
comparisons for the test dataset.
The IsoGCN demonstrably outperforms all other baseline models.
Moreover, owing to the computationally efficient isometric transformation invariant nature of IsoGCNs,
it also achieved a high prediction performance for the meshes that had a significantly
larger graph than those considered in the training dataset.
The IsoGCN can scale up to 1M vertices, which is practical and
is considerably greater than that reported in~\citet{sanchez2020physics}.
Therefore, we conclude that IsoGCN models can be trained on relatively smaller meshes\footnote{However,
it should also be sufficiently large to express sample shapes and fields.}
to save the training time and then used to apply the inference to larger meshes
without observing significant performance deterioration.

Table~\ref{tab:vs_fistr} reports the preprocessing and inference computation time using the equivariant models
with $m = 2$ as the number of hops and FEA using FrontISTR 5.0.0.
We varied the time step ($\Delta t = 1.0, 0.5$) for the FEA computation to compute the $t = 1.0$ time evolution thus, 
resulting in different computation times and errors compared to an FEA with $\Delta t = 0.01$,
which was considered as the ground truth.
Clearly, the IsoGCN is 3- to 5- times faster than the FEA with the same level of accuracy,
while other equivariant models have almost the same speed as FrontISTR with $\Delta t = 0.5$.

\begin{wraptable}{R}{0.6\textwidth}
  \centering
  \caption{Summary of the test losses (mean squared error
  $\pm$ the standard error of the mean in the original scale) of the anisotropic nonlinear heat dataset.
  Here, if ``$\vx$'' is ``Yes'', $\vx$ is also in the input feature.
  We show only the best setting for each method except for the equivariant models.
  For the full table, see Appendix~\ref{sec:differential}.
  OOM denotes the out-of-memory on the applied GPU (32 GiB).}
  \begin{tabular}{lccc}
    \textbf{Method} & \textbf{\# hops} & \boldsymbol{$\vx$}
    & \makecell{\textbf{Loss}\\$\times 10^{-3}$}
    \\
    \hhline{====}
    GIN & 2 & No & 16.921 $\pm$ 0.040 %
    \\
    GCN & 2 & No & 10.427 $\pm$ 0.028 %
    \\
    GCNII & 5 & No & 8.377 $\pm$ 0.024 %
    \\
    Gluster-GCN & 2 & No & 7.266 $\pm$ 0.021 %
    \\
    SGCN & 5 & No & 6.426 $\pm$ 0.018 %
    \\
    \hline
    \multirow{2}{*}{TFN}
    & 2 & No & 15.661 $\pm$ 0.019 %
    \\
    & 5 & No & OOM
    \\[3pt]
    \multirow{2}{*}{SE(3)-Trans.}
    & 2 & No & 14.164 $\pm$ 0.018 %
    \\
    & 5 & No & OOM
    \\[3pt]
    \multirow{2}{*}{\textbf{IsoGCN} (Ours)}
    & 2 & No & 4.674 $\pm$ 0.014 %
    \\
    & 5 & No & \textbf{2.470} $\pm$ 0.008 %
  \end{tabular}
  \label{tab:nl_tensor}
\end{wraptable}

\section{Conclusion}
In this study, we proposed the GCN-based isometric transformation invariant
and equivariant models called \emph{IsoGCN}.
We discussed an example of an isometric adjacency matrix (IsoAM)
that was closely related to the essential differential operators.
The experiment results confirmed that the proposed model leveraged
the spatial structures and can deal with large-scale graphs.
The computation time of the IsoGCN model is significantly shorter than the FEA,
which other equivariant models cannot achieve.
Therefore, IsoGCN must be the first choice to learn physical simulations because of
its computational efficiency as well as isometric transformation invariance and equivariance.
Our demonstrations were conducted on the mesh structured dataset based on the FEA results.
However, we expect IsoGCNs to be applied to various domains, such as object detection,
molecular property prediction, and physical simulations using particles.

\subsubsection*{Acknowledgments}
The authors gratefully acknowledge Takanori Maehara for his helpful advice and NVIDIA for hardware donations.
This work was supported by JSPS KAKENHI Grant Number 19H01098.

\bgroup
\def\arraystretch{1.0}
\begin{table}[H]
  \centering
  \caption{Comparison of computation time.
    To generate the test data, we sampled
    CAD data from the test dataset and then generated the mesh for the graph
    to expand while retaining the element volume at almost the same size.
    The initial temperature field and the material properties
    are set randomly using the same methodology as the dataset sample generation.
    For a fair comparison,
    each computation was run on the same CPU (Intel Xeon E5-2695 v2@2.40GHz)
    using one core, and
    we excluded file I/O time from the measured time.
    OOM denotes the out-of-memory (500 GiB).
    }
  \begin{tabular}{lcccccc}
    & \multicolumn{2}{c}{\boldsymbol{$\vert\V\vert = 21,289$}}
    & \multicolumn{2}{c}{\boldsymbol{$\vert\V\vert = 155,019$}}
    & \multicolumn{2}{c}{\boldsymbol{$\vert\V\vert = 1,011,301$}}
    \\[3pt]
    \textbf{Method}
    & \makecell{\textbf{Loss}\\$\times 10^{-4}$} & \textbf{Time [s]}
    & \makecell{\textbf{Loss}\\$\times 10^{-4}$} & \textbf{Time [s]}
    & \makecell{\textbf{Loss}\\$\times 10^{-4}$} & \textbf{Time [s]}
    \\
    \hhline{=======}
    FrontISTR ($\Delta t = 1.0$) & 10.9 & 16.7 & 6.1 & 181.7 & 2.9 & 1656.5
    \\
    FrontISTR ($\Delta t = 0.5$) & 0.8 & 30.5 & 0.4 & 288.0 & 0.2 & 2884.2
    \\
    TFN & 77.9 & 46.1 & 30.1 & 400.9 & OOM & OOM
    \\
    SE(3)-Transformer & 111.4 & 31.2 & 80.3 & 271.1 & OOM & OOM
    \\
    \textbf{IsoGCN} (Ours) & 8.1 & \textbf{7.4} & 4.9 & \textbf{84.1} & 3.9 & \textbf{648.4}
  \end{tabular}
  \label{tab:vs_fistr}
\end{table}
\egroup

\clearpage

\bibliography{iclr2021_conference}

\begin{thebibliography}{38}
\providecommand{\natexlab}[1]{#1}
\providecommand{\url}[1]{\texttt{#1}}
\expandafter\ifx\csname urlstyle\endcsname\relax
  \providecommand{\doi}[1]{doi: #1}\else
  \providecommand{\doi}{doi: \begingroup \urlstyle{rm}\Url}\fi

\bibitem[Ahmed et~al.(2018)Ahmed, Saint, Shabayek, Cherenkova, Das, Gusev,
  Aouada, and Ottersten]{ahmed2018survey}
Eman Ahmed, Alexandre Saint, Abd El~Rahman Shabayek, Kseniya Cherenkova, Rig
  Das, Gleb Gusev, Djamila Aouada, and Bjorn Ottersten.
\newblock A survey on deep learning advances on different 3d data
  representations.
\newblock \emph{arXiv preprint arXiv:1808.01462}, 2018.

\bibitem[Alet et~al.(2019)Alet, Jeewajee, Villalonga, Rodriguez, Lozano-Perez,
  and Kaelbling]{alet19a}
Ferran Alet, Adarsh~Keshav Jeewajee, Maria~Bauza Villalonga, Alberto Rodriguez,
  Tomas Lozano-Perez, and Leslie Kaelbling.
\newblock Graph element networks: adaptive, structured computation and memory.
\newblock In \emph{ICML}, 2019.

\bibitem[Baskin et~al.(1997)Baskin, Palyulin, and Zefirov]{baskin1997neural}
Igor~I Baskin, Vladimir~A Palyulin, and Nikolai~S Zefirov.
\newblock A neural device for searching direct correlations between structures
  and properties of chemical compounds.
\newblock \emph{Journal of chemical information and computer sciences},
  37\penalty0 (4):\penalty0 715--721, 1997.

\bibitem[Battaglia et~al.(2018)Battaglia, Hamrick, Bapst, Sanchez-Gonzalez,
  Zambaldi, Malinowski, Tacchetti, Raposo, Santoro, Faulkner,
  et~al.]{battaglia2018relational}
Peter~W Battaglia, Jessica~B Hamrick, Victor Bapst, Alvaro Sanchez-Gonzalez,
  Vinicius Zambaldi, Mateusz Malinowski, Andrea Tacchetti, David Raposo, Adam
  Santoro, Ryan Faulkner, et~al.
\newblock Relational inductive biases, deep learning, and graph networks.
\newblock \emph{arXiv preprint arXiv:1806.01261}, 2018.

\bibitem[Chang \& Cheng(2020)Chang and Cheng]{chang2020learning}
Kai-Hung Chang and Chin-Yi Cheng.
\newblock Learning to simulate and design for structural engineering.
\newblock \emph{arXiv preprint arXiv:2003.09103}, 2020.

\bibitem[Chen et~al.(2020)Chen, Wei, Huang, Ding, and Li]{chen2020simple}
Ming Chen, Zhewei Wei, Zengfeng Huang, Bolin Ding, and Yaliang Li.
\newblock Simple and deep graph convolutional networks.
\newblock \emph{arXiv preprint arXiv:2007.02133}, 2020.

\bibitem[Chiang et~al.(2019)Chiang, Liu, Si, Li, Bengio, and
  Hsieh]{chiang2019cluster}
Wei-Lin Chiang, Xuanqing Liu, Si~Si, Yang Li, Samy Bengio, and Cho-Jui Hsieh.
\newblock Cluster-gcn: An efficient algorithm for training deep and large graph
  convolutional networks.
\newblock In \emph{Proceedings of the 25th ACM SIGKDD International Conference
  on Knowledge Discovery \& Data Mining}, pp.\  257--266, 2019.

\bibitem[Cohen \& Welling(2016)Cohen and Welling]{cohen2016group}
Taco Cohen and Max Welling.
\newblock Group equivariant convolutional networks.
\newblock In \emph{International conference on machine learning}, pp.\
  2990--2999, 2016.

\bibitem[Cohen et~al.(2018)Cohen, Geiger, K{\"o}hler, and
  Welling]{cohen2018spherical}
Taco~S Cohen, Mario Geiger, Jonas K{\"o}hler, and Max Welling.
\newblock Spherical cnns.
\newblock In \emph{ICLR}, 2018.

\bibitem[Cohen et~al.(2019)Cohen, Weiler, Kicanaoglu, and
  Welling]{cohen2019gauge}
Taco~S Cohen, Maurice Weiler, Berkay Kicanaoglu, and Max Welling.
\newblock Gauge equivariant convolutional networks and the icosahedral cnn.
\newblock \emph{ICML}, 2019.

\bibitem[Fey \& Lenssen(2019)Fey and Lenssen]{Fey/Lenssen/2019}
Matthias Fey and Jan~E. Lenssen.
\newblock Fast graph representation learning with {PyTorch Geometric}.
\newblock In \emph{ICLR Workshop on Representation Learning on Graphs and
  Manifolds}, 2019.

\bibitem[Fuchs et~al.(2020)Fuchs, Worrall, Fischer, and Welling]{fuchs2020se}
Fabian Fuchs, Daniel Worrall, Volker Fischer, and Max Welling.
\newblock Se (3)-transformers: 3d roto-translation equivariant attention
  networks.
\newblock \emph{Advances in Neural Information Processing Systems}, 33, 2020.

\bibitem[Geuzaine \& Remacle(2009)Geuzaine and Remacle]{gauzaine2009}
Christophe Geuzaine and Jean-Fran\c{c}ois Remacle.
\newblock Gmsh: a three-dimensional finite element mesh generator with built-in
  pre- and post-processing facilities.
\newblock \emph{International Journal for Numerical Methods in Engineering},
  79\penalty0 (11):\penalty0 1309--1331, 2009.

\bibitem[Gilmer et~al.(2017)Gilmer, Schoenholz, Riley, Vinyals, and
  Dahl]{gilmer2017neural}
Justin Gilmer, Samuel~S Schoenholz, Patrick~F Riley, Oriol Vinyals, and
  George~E Dahl.
\newblock Neural message passing for quantum chemistry.
\newblock In \emph{Proceedings of the 34th International Conference on Machine
  Learning-Volume 70}, pp.\  1263--1272. JMLR. org, 2017.

\bibitem[Gori et~al.(2005)Gori, Monfardini, and Scarselli]{gori2005new}
Marco Gori, Gabriele Monfardini, and Franco Scarselli.
\newblock A new model for learning in graph domains.
\newblock In \emph{Proceedings. 2005 IEEE International Joint Conference on
  Neural Networks, 2005.}, volume~2, pp.\  729--734. IEEE, 2005.

\bibitem[He et~al.(2016)He, Zhang, Ren, and Sun]{he2016deep}
Kaiming He, Xiangyu Zhang, Shaoqing Ren, and Jian Sun.
\newblock Deep residual learning for image recognition.
\newblock In \emph{Proceedings of the IEEE conference on computer vision and
  pattern recognition}, pp.\  770--778, 2016.

\bibitem[Ihara et~al.(2017)Ihara, Hashimoto, and Okuda]{ihara2017web}
Yu~Ihara, Gaku Hashimoto, and Hiroshi Okuda.
\newblock Web-based integrated cloud cae platform for large-scale finite
  element analysis.
\newblock \emph{Mechanical Engineering Letters}, 3:\penalty0 17--00520, 2017.

\bibitem[Kingma \& Ba(2014)Kingma and Ba]{kingma2014adam}
Diederik~P Kingma and Jimmy Ba.
\newblock Adam: A method for stochastic optimization.
\newblock \emph{arXiv preprint arXiv:1412.6980}, 2014.

\bibitem[Kipf \& Welling(2017)Kipf and Welling]{kipf2017semi}
Thomas~N Kipf and Max Welling.
\newblock Semi-supervised classification with graph convolutional networks.
\newblock In \emph{ICLR}, 2017.

\bibitem[Klicpera et~al.(2020)Klicpera, Gro{\ss}, and
  G{\"u}nnemann]{klicpera_dimenet_2020}
Johannes Klicpera, Janek Gro{\ss}, and Stephan G{\"u}nnemann.
\newblock Directional message passing for molecular graphs.
\newblock In \emph{ICLR}, 2020.

\bibitem[Koch et~al.(2019)Koch, Matveev, Jiang, Williams, Artemov, Burnaev,
  Alexa, Zorin, and Panozzo]{Koch_2019_CVPR}
Sebastian Koch, Albert Matveev, Zhongshi Jiang, Francis Williams, Alexey
  Artemov, Evgeny Burnaev, Marc Alexa, Denis Zorin, and Daniele Panozzo.
\newblock Abc: A big cad model dataset for geometric deep learning.
\newblock In \emph{The IEEE Conference on Computer Vision and Pattern
  Recognition (CVPR)}, June 2019.

\bibitem[Kondor(2018)]{kondor2018n}
Risi Kondor.
\newblock N-body networks: a covariant hierarchical neural network architecture
  for learning atomic potentials.
\newblock \emph{arXiv preprint arXiv:1803.01588}, 2018.

\bibitem[Maron et~al.(2018)Maron, Ben-Hamu, Shamir, and
  Lipman]{maron2018invariant}
Haggai Maron, Heli Ben-Hamu, Nadav Shamir, and Yaron Lipman.
\newblock Invariant and equivariant graph networks.
\newblock \emph{arXiv preprint arXiv:1812.09902}, 2018.

\bibitem[Morita et~al.(2016)Morita, Yonekura, Yasuzumi, Tsunori, Hashimoto, and
  Okuda]{morita2016development}
Naoki Morita, Kazuo Yonekura, Ichiro Yasuzumi, Mitsuyoshi Tsunori, Gaku
  Hashimoto, and Hiroshi Okuda.
\newblock Development of 3$\times$ 3 dof blocking structural elements to
  enhance the computational intensity of iterative linear solver.
\newblock \emph{Mechanical Engineering Letters}, 2:\penalty0 16--00082, 2016.

\bibitem[Nair \& Hinton(2010)Nair and Hinton]{nair2010rectified}
Vinod Nair and Geoffrey~E Hinton.
\newblock Rectified linear units improve restricted boltzmann machines.
\newblock In \emph{Proceedings of the 27th international conference on machine
  learning (ICML-10)}, pp.\  807--814, 2010.

\bibitem[Paszke et~al.(2019)Paszke, Gross, Massa, Lerer, Bradbury, Chanan,
  Killeen, Lin, Gimelshein, Antiga, Desmaison, Kopf, Yang, DeVito, Raison,
  Tejani, Chilamkurthy, Steiner, Fang, Bai, and Chintala]{NEURIPS2019_9015}
Adam Paszke, Sam Gross, Francisco Massa, Adam Lerer, James Bradbury, Gregory
  Chanan, Trevor Killeen, Zeming Lin, Natalia Gimelshein, Luca Antiga, Alban
  Desmaison, Andreas Kopf, Edward Yang, Zachary DeVito, Martin Raison, Alykhan
  Tejani, Sasank Chilamkurthy, Benoit Steiner, Lu~Fang, Junjie Bai, and Soumith
  Chintala.
\newblock Pytorch: An imperative style, high-performance deep learning library.
\newblock In H.~Wallach, H.~Larochelle, A.~Beygelzimer, F.~d\textquotesingle
  Alch\'{e}-Buc, E.~Fox, and R.~Garnett (eds.), \emph{Advances in Neural
  Information Processing Systems 32}, pp.\  8024--8035. Curran Associates,
  Inc., 2019.

\bibitem[Sanchez-Gonzalez et~al.(2018)Sanchez-Gonzalez, Heess, Springenberg,
  Merel, Riedmiller, Hadsell, and Battaglia]{sanchez2018graph}
Alvaro Sanchez-Gonzalez, Nicolas Heess, Jost~Tobias Springenberg, Josh Merel,
  Martin Riedmiller, Raia Hadsell, and Peter Battaglia.
\newblock Graph networks as learnable physics engines for inference and
  control.
\newblock \emph{arXiv preprint arXiv:1806.01242}, 2018.

\bibitem[Sanchez-Gonzalez et~al.(2019)Sanchez-Gonzalez, Bapst, Cranmer, and
  Battaglia]{sanchez2019hamiltonian}
Alvaro Sanchez-Gonzalez, Victor Bapst, Kyle Cranmer, and Peter Battaglia.
\newblock Hamiltonian graph networks with ode integrators.
\newblock \emph{arXiv preprint arXiv:1909.12790}, 2019.

\bibitem[Sanchez-Gonzalez et~al.(2020)Sanchez-Gonzalez, Godwin, Pfaff, Ying,
  Leskovec, and Battaglia]{sanchez2020physics}
Alvaro Sanchez-Gonzalez, Jonathan Godwin, Tobias Pfaff, Rex Ying, Jure
  Leskovec, and Peter~W Battaglia.
\newblock Learning to simulate complex physics with graph networks.
\newblock \emph{arXiv preprint arXiv:2002.09405}, 2020.

\bibitem[Scarselli et~al.(2008)Scarselli, Gori, Tsoi, Hagenbuchner, and
  Monfardini]{scarselli2008graph}
Franco Scarselli, Marco Gori, Ah~Chung Tsoi, Markus Hagenbuchner, and Gabriele
  Monfardini.
\newblock The graph neural network model.
\newblock \emph{IEEE Transactions on Neural Networks}, 20\penalty0
  (1):\penalty0 61--80, 2008.

\bibitem[Sperduti \& Starita(1997)Sperduti and Starita]{sperduti1997supervised}
Alessandro Sperduti and Antonina Starita.
\newblock Supervised neural networks for the classification of structures.
\newblock \emph{IEEE Transactions on Neural Networks}, 8\penalty0 (3):\penalty0
  714--735, 1997.

\bibitem[Swartz \& Wendroff(1969)Swartz and Wendroff]{swartz1969generalized}
Blair Swartz and Burton Wendroff.
\newblock Generalized finite-difference schemes.
\newblock \emph{Mathematics of Computation}, 23\penalty0 (105):\penalty0
  37--49, 1969.

\bibitem[Tamai \& Koshizuka(2014)Tamai and Koshizuka]{tamai2014least}
Tasuku Tamai and Seiichi Koshizuka.
\newblock Least squares moving particle semi-implicit method.
\newblock \emph{Computational Particle Mechanics}, 1\penalty0 (3):\penalty0
  277--305, 2014.

\bibitem[Thomas et~al.(2018)Thomas, Smidt, Kearnes, Yang, Li, Kohlhoff, and
  Riley]{thomas2018tensor}
Nathaniel Thomas, Tess Smidt, Steven Kearnes, Lusann Yang, Li~Li, Kai Kohlhoff,
  and Patrick Riley.
\newblock Tensor field networks: Rotation-and translation-equivariant neural
  networks for 3d point clouds.
\newblock \emph{arXiv preprint arXiv:1802.08219}, 2018.

\bibitem[Weiler et~al.(2018)Weiler, Geiger, Welling, Boomsma, and
  Cohen]{weiler20183d}
Maurice Weiler, Mario Geiger, Max Welling, Wouter Boomsma, and Taco~S Cohen.
\newblock 3d steerable cnns: Learning rotationally equivariant features in
  volumetric data.
\newblock In \emph{NeurIPS}, pp.\  10381--10392, 2018.

\bibitem[Wu et~al.(2019)Wu, Souza, Zhang, Fifty, Yu, and
  Weinberger]{pmlr-v97-wu19e}
Felix Wu, Amauri Souza, Tianyi Zhang, Christopher Fifty, Tao Yu, and Kilian
  Weinberger.
\newblock Simplifying graph convolutional networks.
\newblock In \emph{ICML}, pp.\  6861--6871. PMLR, 2019.

\bibitem[Xu et~al.(2018)Xu, Hu, Leskovec, and Jegelka]{xu2018powerful}
Keyulu Xu, Weihua Hu, Jure Leskovec, and Stefanie Jegelka.
\newblock How powerful are graph neural networks?
\newblock \emph{arXiv preprint arXiv:1810.00826}, 2018.

\bibitem[You et~al.(2019)You, Ying, and Leskovec]{you2019position}
Jiaxuan You, Rex Ying, and Jure Leskovec.
\newblock Position-aware graph neural networks.
\newblock \emph{arXiv preprint arXiv:1906.04817}, 2019.

\end{thebibliography}
\bibliographystyle{iclr2021_conference}

\appendix
\section{Notation}
\bgroup
\def\arraystretch{1.5}
\begin{tabular}{p{0.2\textwidth}p{0.7\textwidth}}
  $\displaystyle \G$ & A graph
  \\
  $\displaystyle \V$ & A vertex set
  \\
  $\displaystyle \VV$ & The number of vertices
  \\
  $\displaystyle \E$ & An edge set
  \\
  $\displaystyle \Z^+$ & The positive integers
  \\
  $\displaystyle d$ & The dimension of the Euclidean space
  \\
  $\displaystyle \vx_i$ & The position of the $i$th vertex
  \\
  $\displaystyle \evx_{ik}$ & Element $k$ of $\vx_i$
  \\
  $\displaystyle \tG \in \R^{\VV \times \VV \times d}$ & The isometric adjacency matrix (IsoAM) (\eqref{eq:isoam})
  \\
  $\displaystyle \tG_{ij;;:} \in \R^{d}$ & Slice of $\tG$ in the spatial index (\eqref{eq:isoam})
  \\
  $\displaystyle \tG_{ij;;k} \in \R$ & Element $(i, j, k)$ of $\tG$
  \\
  $\displaystyle \tH\supp{p} \in \R^{\VV \times f \times d^p}$ & A rank-$p$ tensor field tensor ($f, p \in \Z^+$)
  \\
  $\displaystyle \etH\supp{p}_{i;g;k_1 k_2 \dots k_p}$ & Element $(i; g; k_1, k_2, \dots, k_p)$ of $\tH\supp{p}$.
  $i$ refers to the permutation representation,
  $k_1, \dots k_p$ refer to the Euclidean representation, and $g$ denotes
  the feature index (See section~\ref{sec:model}).
  \\
  $\displaystyle \left[ \bigotimes^p \tG \right] \ast \tH\supp{0}$
  & Convolution of the $p$th power of $\tG$ and rank-0 tensor field $\tH\supp{0}$
  (\eqref{eq:convolution1}, \eqref{eq:convolution_p})
  \\
  $\displaystyle \left[ \bigotimes^p \tG \right] \odot \tH\supp{q}$
  & Contraction of the $p$th power of $\tG$ and rank-$q$ tensor fields
  (\eqref{eq:contraction_1}, \eqref{eq:contraction_p})
  \\
  $\displaystyle \left[ \bigotimes^p \tG \right] \otimes \tH\supp{q}$
  & Tensor product of the $p$th power of $\tG$ and rank-$q$ tensor fields $\tH\supp{q}$ (\eqref{eq:tp_pq})
  \\
  $\displaystyle \tH\sub{in}\supp{p}$ & The rank-$p$ input tensor field of the considered layer
  \\
  $\displaystyle \tH\sub{out}\supp{p}$ & The rank-$p$ output tensor field of the considered layer
  \\
  $\displaystyle \sigma$ & The activation function
  \\
  $\displaystyle \mW$ & The trainable parameter matrix
  \\
  $\displaystyle \mA \in \mathbb{R}^{\VV \times \VV}$ & An adjacency matrix
  \\
  $\displaystyle \delta_{ij}$ & The Kronecker delta
  \\
  $\displaystyle V_i\sup{effective}$ & The effective volume of the $i$th vertex (\eqref{eq:v_effective})
  \\
  $\displaystyle V_i\sup{mean}$ & The mean volume of the $i$th vertex (\eqref{eq:v_mean})
  \\
  $\displaystyle \tilde{\tD} \in \R^{\VV \times \VV}$ & A concrete instance of IsoAM (\eqref{eq:tilde_d})
  \\
\end{tabular}
\egroup

\section{Proofs of propositions\label{sec:proofs}}
In this section, we present the proofs of the propositions described in Section~\ref{sec:model}.
Let
$\R^3 \ni \vg(\vx_l, \vx_k) = (\vx_k - \vx_l)$.
Note that $\tG$ is expressed using $\vg(\vx_i, \vx_j)$ as
$\tG_{ij;;:} = \sum_{k, l \in \V, k \neq l} \mT_{ijkl} \vg(\vx_l, \vx_k)$.

\subsection{Proof of Proposition~\ref{prop:gam_iso}}
\begin{proof}
  First, we demonstrate the invariance with respect to the translation with $\forall\vt \in \R^d$.
  $\vg(\vx_i, \vx_j)$ is transformed invariantly as follows under translation:
  \begin{align}
    \vg(\vx_i + \vt, \vx_j + \vt) &= [\vx_j + \vt - (\vx_i + \vt)]
    \nonumber
    \\
    &= (\vx_j - \vx_i)
    \nonumber
    \\
    &= \vg(\vx_i, \vx_j).
  \end{align}
  By definition, $\mT_{ijkl}$ is also translation invariant.
  Thus,
  \begin{align}
    \sum_{k, l \in \V, k \neq l}\mT_{ijkl} \vg(\vx_l + \vt, \vx_k + \vt)
    &= \sum_{k, l \in \V, k \neq l}\mT_{ijkl} \vg(\vx_l, \vx_k)
    \nonumber
    \\
    &= \tG_{ij;;:}.
  \end{align}
  We then show an equivariance regarding the orthogonal transformation with $\forall \mU \in \mathrm{O}(d)$.
  $\vg(\vx_i, \vx_j)$ is transformed as follows by orthogonal transformation:
  \begin{align}
    \vg(\mU \vx_i, \mU \vx_j) &= \mU \vx_j- \mU \vx_i
    \nonumber
    \\
    &= \mU (\vx_j - \vx_i)
    \nonumber
    \\
    &= \mU \vg(\vx_i, \vx_j).
  \end{align}
  By definition, $\mT_{ijkl}$ is transformed to $\mU\mT_{ijkl}\mU^{-1}$
  by orthogonal transformation. Thus,
  \begin{align}
    \sum_{k, l \in \V, k \neq l} \mU\mT_{ijkl}\mU^{-1} \vg(\mU \vx_l, \mU \vx_k)
    &= \sum_{k, l \in \V, k \neq l} \mU\mT_{ijkl}\mU^{-1} \mU \vg(\vx_l, \vx_k)
    \nonumber
    \\
    &= \mU \tG_{ij;;:}.
  \end{align}
  Therefore, $\tG$ is both translation invariant and an orthogonal transformation
  equivariant.
\end{proof}

\subsection{Proof of Proposition~\ref{prop:gam_contraction}}
\begin{proof}
  Here, $\tG \odot \tG$ is translation invariant because
  $\tG$ is translation invariant.
  We prove rotation invariance
  under an orthogonal transformation $\forall \mU \in \O(n)$.
  In addition, $\tG \odot \tG$ is transformed under $\mU$ as follows:
  \begin{align}
    \sum_{j, k} \etG_{ij;;k} \etG_{jl;;k}
    &\mapsto \sum_{j, k, m, n} \mU_{km} \etG_{ij;;m} \mU_{kn} \etG_{jl;;n}
    \nonumber
    \\
    &= \sum_{j, k, m, n} \mU_{km} \mU_{kn} \etG_{ij;;m} \etG_{jl;;n}&
    \nonumber
    \\
    &= \sum_{j, k, m, n} \mU^T_{mk} \mU_{kn} \etG_{ij;;m} \etG_{jl;;n}&
    \nonumber
    \\
    &= \sum_{j, m, n} \delta_{mn} \etG_{ij;;m} \etG_{jl;;n}\ \ &(\because \text{property of the orthogonal matrix})
    \nonumber
    \\
    &= \sum_{j} \etG_{ij;;m} \etG_{jl;;m}&
    \nonumber
    \\
    &= \sum_{j, k} \etG_{ij;;k} \etG_{jl;;k}.& (\because \text{Change the dummy index $m \to k$})
  \end{align}
  Therefore, $\tG \odot \tG$ is isometric transformation invariant.
\end{proof}

\subsection{Proof of Proposition~\ref{prop:gam_tensor}}
\begin{proof}
  $\tG \otimes \tG$ is transformed under $\forall \mU \in \O(n)$ as follows:
  \begin{align}
    \sum_{j} \etG_{ij;;k} \etG_{jl;;m}
    &\mapsto \sum_{n, o} \mU_{kn} \etG_{ij;;n} \mU_{mo} \etG_{jl;;o}
    \nonumber
    \\
    &= \sum_{n, o} \mU_{kn} \etG_{ij;;n} \etG_{jl;;o} \mU_{om}^T.
  \end{align}
  By regarding $\etG_{ij;;n} \etG_{jl;;o}$ as one matrix $\emH_{no}$, it follows
  the coordinate transformation of rank-2 tensor $\mU \mH \mU^T$ for each $i$, $j$, and $l$.
\end{proof}

\section{Physical intuition of \texorpdfstring{$\tilde{\tD}$}{D}}\label{sec:operators}
In this section, we discuss the connection between the concrete IsoAM example 
$\tilde{\tD}$ and the differential operators such as the gradient, divergence,
the Laplacian, the Jacobian, and the Hessian operators.

Let
$\phi_i \in \R$
denote a rank-0 tensor (scalar) at the $i$th vertex.
Let us assume a partial derivative model of a rank-0 tensor $\phi$ at the $i$th vertex regarding the $k$th axis
$(\inpdiff{\phi}{x_k})_i \in \R \ (k \in \{1, \dots, d\})$,
that is based on the gradient model in the least squares moving particle
semi-implicit method \citep{tamai2014least}.
\begin{align}
  {\left(\pdiff{\phi}{x_k}\right)}_i
  :=& \mM^{-1}_i \sum_j \frac{\phi_j - \phi_i}{\Vert\vx_{j} - \vx_{i}\Vert}
  \frac{\evx_{jk} - \evx_{ik}}{\lVert \vx_{j} - \vx_{i} \rVert}
  w_{ij} \emA_{ij}(m)
  \\
  =& \sum_j \etD_{ijk} (\phi_j - \phi_i),
  \label{eq:pdiff}
  \\
  \mM_i =& \sum_l \frac{\vx_l - \vx_i}{\lVert \vx_{l} - \vx_{i} \rVert}
  \otimes \frac{\vx_l - \vx_i}{\lVert \vx_l - \vx_i \rVert} w_{il} \emA_{il}(m).
\end{align}
Although one could define $w_{ij}$ as a function of the distance
$\Vert \vx_j - \vx_i \Vert$,
$w_{ij}$ was kept constant with respect to the distance required to maintain the simplicity of the model
with fewer hyperparameters.

\subsection{Gradient}
$\tilde{\tD}$ can be viewed as a Laplacian matrix based on $\tD$;
however, $\tilde{\tD} \ast \tH\supp{0}$ can be interpreted as the gradient
within the Euclidean space.
Let
$\nabla \ \tH\supp{0} \in \R^{\VV \times f \times d}$
be an approximation of the gradient of $\tH\supp{0}$.
Using \eqref{eq:pdiff}, the gradient model can be expressed as follows:
\begin{align}
  {\left( \nabla \ \tH\supp{0} \right)}_{i;g;k}
  &= \pdiff{\etH\supp{0}_{i;g;}}{x_k}
  \\
  &= \etD_{ijk} (\etH\supp{0}_{j;g;} - \etH\supp{0}_{i;g;}).
\end{align}
Using this gradient model, we can confirm that
${(\tilde{\tD} \ast \tH\supp{0})}_{i;g;k} = {(\nabla \ \tH\supp{0})}_{i;glk}$
because
\begin{align}
  {\left( \tilde{\tD} \ast \tH\supp{0} \right)}_{i;g;k}
  &= \sum_j \tilde{\etD}_{ij;;k} \etH\supp{0}_{j;g;}
  \\
  &= \sum_j (\etD_{ij;;k} - \delta_{ij} \sum_l \etD_{il;;k}) \etH\supp{0}_{j;g;}&
  \nonumber
  \\
  &= \sum_j \etD_{ij;;k} \etH\supp{0}_{j;g;} - \sum_{j, l} \delta_{ij} \etD_{il;;k} \etH\supp{0}_{j;g;}&
  \nonumber
  \\
  &= \sum_j \etD_{ij;;k} \etH\supp{0}_{j;g;} - \sum_{l} \etD_{il;;k} \etH\supp{0}_{i;g;}&
  \nonumber
  \\
  &= \sum_j \etD_{ij;;k} \etH\supp{0}_{j;g;} - \sum_{j} \etD_{ij;;k} \etH\supp{0}_{i;g;}& (\because \text{Change the dummy index $l \to j$})
  \nonumber
  \\
  &= \sum_j \etD_{ij;;k} (\etH\supp{0}_{j;g;} - \etH\supp{0}_{i;g;})&
  \nonumber
  \\
  &= {\left( \nabla \ \tH\supp{0} \right)}_{i;g;k}.
\end{align}
Therefore, $\tilde{\tD} \ast$ can be interpreted as the gradient operator within a
Euclidean space.

\subsection{Divergence}
We show that $\tilde{\tD} \odot \tH\supp{1}$ corresponds to the divergence.
Using $\tD$, the divergence model
$\nabla \cdot \tH\supp{1} \in \R^{\VV \times f}$
is expressed as follows:
\begin{align}
  {\left( \nabla \cdot \tH\supp{1} \right)}_{i;g;}
  &= {\left( \sum_k \pdiff{\ \tH\supp{1}}{x_k} \right)}_{i;g;}
  \\
  &= \sum_{j,k}\etD_{ij;;k} (\etH\supp{1}_{j;g;k} - \etH\supp{1}_{i;g;k}).
\end{align}

Then, $\tilde{\tD} \odot \tH\supp{1}$ is
\begin{align}
  {(\tilde{\tD} \odot \tH\supp{1})}_{i;g;}
  &= \sum_{j,k} \tilde{\tD}_{ij;;k} \etH\supp{1}_{i;g;k}&
  \nonumber
  \\
  &= \sum_{j,k} \left( \tD_{ij;;k} - \delta_{ij} \sum_l \tD \right) \ \etH\supp{1}_{i;g;k}&
  \nonumber
  \\
  &= \sum_{j,k} \tD_{ij;;k} \tH\supp{1}_{j;g;k} - \sum_{l,k} \tD_{il;;k} \tH\supp{1}_{i;g;k}&
  \nonumber
  \\
  &= \sum_{j,k}\etD_{ij;;k} (\etH\supp{1}_{j;g;k} - \etH\supp{1}_{i;g;k})
  & (\because \text{Change the dummy index $l \to j$})
  \nonumber
  \\
  &= {(\nabla \cdot \tH\supp{1})}_{i;g;}.
\end{align}

\subsection{Laplacian operator}
We prove that $\tilde{\tD} \odot \tilde{\tD}$ corresponds to the Laplacian
operator within a Euclidean space.

Using \eqref{eq:pdiff}, the Laplacian model
$\nabla^2 \ \tH\supp{0} \in \R^{\VV \times f}$
can be expressed as follows:
\begin{align}
  {\left( \nabla^2 \ \tH\supp{0} \right)}_{i;g;}
  :=& \sum_k {\left[ \frac{\partial}{\partial x_k} {\left( \frac{\partial \tH}{\partial x_k} \right)}_i \right]}_{i;g;}
  \nonumber
  \\
  =& \sum_{j, k} \etD_{ij;;k} \left[ {\left( \frac{\partial \tH}{\partial x_k} \right)}_{j;g;}
  - {\left( \frac{\partial \tH}{\partial x_k} \right)}_{i;g;} \right]
  \nonumber
  \\
  =& \sum_{j, k} \etD_{ij;;k} \left[ \sum_l \etD_{jl;;k} (\etH\supp{0}_{l;g;} - \etH\supp{0}_{j;g;})
  - \sum_l \etD_{il;;k} (\etH\supp{0}_{l;g;} - \etH\supp{0}_{i;g;}) \right]
  \nonumber
  \\
  =& \sum_{j, k, l} \etD_{ij;;k} ( \etD_{jl;;k} - \etD_{il;;k} ) (\etH\supp{0}_{l;g;} - \etH\supp{0}_{j;g;}).
\end{align}
Then, $(\tilde{\tD} \odot \tilde{\tD}) \tH\supp{0}$ is
\begin{align}
  {((\tilde{\tD} \odot \tilde{\tD}) \tH\supp{0})}_{i;g;}
  &= \sum_{j, k, l} \tilde{\etD}_{ij;;k} \tilde{\etD}_{jl;;k} \etH\supp{0}_{l;g;}
  \nonumber
  \\
  &= \sum_{j, k, l} \left( \etD_{ij;;k} - \delta_{ij} \sum_m \etD_{im;;k} \right) \left( \etD_{jl;;k} - \delta_{jl} \sum_n \etD_{jn;;k} \right) \etH\supp{0}_{l;g;}&
  \nonumber
  \\
  &=\sum_{j, k, l} \etD_{ij;;k} \etD_{jl;;k} \etH\supp{0}_{l;g;} - \sum_{j, k, n} \etD_{ij;;k} \etD_{jn;;k} \etH\supp{0}_{j;g;}
  \nonumber
  \\
  & \ \ \ \ \ \ \ \ - \sum_{k, l, m} \etD_{im;;k} \etD_{il;;k} \etH\supp{0}_{l;g;} + \sum_{k, m, n} \etD_{im;;k} \etD_{in;;k} \etH\supp{0}_{i;g;}&
  \nonumber
  \\
  &=\sum_{j, k, l} \etD_{ij;;k} \etD_{jl;;k} \etH\supp{0}_{l;g;} - \sum_{j, k, n} \etD_{ij;;k} \etD_{jn;;k} \etH\supp{0}_{j;g;}
  \nonumber
  \\
  & \ \ \ \ \ \ \ \ - \sum_{k, l, j} \etD_{ij;;k} \etD_{il;;k} \etH\supp{0}_{l;g;} + \sum_{k, j, n} \etD_{ij;;k} \etD_{in;;k} \etH\supp{0}_{i;g;}&
  \nonumber
  \\
  & \ \ \ \ \ \ \ \ \ \ \ \ \ \ \ \ \ \ (\because \text{Change the dummy index $m \to j$ for the third and fourth terms})
  \nonumber
  \\
  \nonumber
  \\
  &= \sum_{j, k, l} \etD_{ij;;k} (\etD_{jl;;k} -\etD_{il;;k}) (\etH\supp{0}_{l;g;} - \etH\supp{0}_{j;g;}) &
  \nonumber
  \\
  & \ \ \ \ \ \ \ \ \ \ \ \ \ \ \ \ \ \ (\because \text{Change the dummy index $n \to l$ for the second and fourth terms})
  \nonumber
  \\
  \nonumber
  \\
  &= {\left( \nabla^2 \ \tH\supp{0} \right)}_{i;g;}. &
\end{align}

\subsection{Jacobian and Hessian operators}
Considering a similar discussion, we can show the following dependencies.
For the Jacobian model,
$\mJ[\tH\supp{1}] \in \R^{\VV \times f \times d \times d}$,
\begin{align}
  {\left( \mJ[\tH\supp{1}] \right)}_{i;g;kl}
  &= {\left(\pdiff{\ \tH\supp{1}}{x_l} \right)}_{i;g;k}
  \\
  &= \sum_j \etD_{ij;;l} ( \etH\supp{1}_{j;g;k} - \etH\supp{1}_{i;g;k} )
  \\
  &= {(\tilde{D} \otimes \tH\supp{1})}_{i;g;lk}.
\end{align}

For the Hessian model,
$\mathrm{Hess}[\tH\supp{0}] \in \R^{\VV \times f \times d \times d}$,
\begin{align}
  {\left( \mathrm{Hess}[\tH\supp{0}] \right)}_{i;g;kl}
  &= {\left( \pdiff{}{x_k} \pdiff{}{x_l} \ \tH\supp{0} \right)}_{i;g;}
  \\
  &= \sum_{j, m} \etD_{ij;;k} [
    \etD_{jm;;l} ( \etH\supp{0}_{m;g;} - \etH\supp{0}_{l;g;})
    - \etD_{im;;l} ( \etH\supp{0}_{m;g;} - \etH\supp{0}_{i;g;})]
  \\
  &= {\left[ (\tilde{\tD} \otimes \tilde{\tD}) \ast \ \tH\supp{0} \right]}_{i;g;kl}.
\end{align}

\section{IsoGCN modeling details}\label{sec:modeling}
To achieve isometric transformation invariance and equivariance, there are
several rules to follow.
Here, we describe the desired focus when constructing an IsoGCN model.
In this section, a rank-$p$ tensor denotes
a tensor the rank of which is $p \geq 1$
and $\sigma$ denotes a nonlinear activation function.
$\mW$ is a trainable weight matrix and $\vb$ is a trainable bias.

\subsection{Activation and bias}\label{sec:activation_bias}
As the nonlinear activation function is not isometric transformation equivariant,
nonlinear activation to rank-$p$ tensors cannot be applied,
while one can apply any activation to rank-0 tensors.
In addition, adding bias is also not isometric transformation equivariant,
one cannot add bias when performing an affine transformation to rank-$p$ tensors.
Again, one can add bias to rank-0 tensors.

Thus, for instance, if one converts from rank-0 tensors
$\tH\supp{0}$ to rank-1 tensors
using IsoAM $\tG$, $\tG \ast \sigma(\tH\supp{0} \mW + \vb)$ and
$(\tG \ast \sigma(\tH\supp{0})) \mW$
are isometric equivariant functions, however
$(\tG \ast \tH\supp{0}) \mW + \vb$ and
$\sigma\left((\tG \ast \sigma(\tH\supp{0})) \mW\right)$
are not due to the bias and the nonlinear activation, respectively.
Likewise, regarding a conversion from rank-1 tensors $ \tH\supp{1}$ to rank-0
tensors,
$\sigma\left((\tG \odot \tH\supp{1}) \mW + \vb\right)$ and
$\sigma\left(\tG \odot (\tH\supp{1} \mW)\right)$ are isometric transformation invariant functions; however, 
$\tG \odot (\tH\supp{1} \mW + \vb)$ and
$(\tG \odot \sigma(\tH\supp{1})) \mW + \vb$ are not.

To convert rank-$p$ tensors to rank-$q$ tensors ($q \geq 1$),
one can apply neither bias nor nonlinear activation.
To add nonlinearity to such a conversion, we can multiply the converted rank-0 tensors
$\sigma((\bigotimes^p \tG \odot \tH\supp{p}) \mW + \vb)$
with the input tensors $\tH\supp{p}$ or the output tensors $\tH\supp{q}$.

\subsection{Preprocessing of input feature}
Similarly to the discussion regarding the biases,
we have to take care of the preprocessing of rank-$p$ tensors to retain
isometric transformation invariance because adding a constant array and
component-wise scaling could distort the tensors, resulting in broken isometric
transformation equivariance.

For instance, $\tH\supp{p} / \mathrm{Std}_\mathrm{all}\left[\tH\supp{p}\right]$ is a valid transformation to
retain isometric transformation equivariance, assuming
$\mathrm{Std}_\mathrm{all}\left[\tH\supp{p}\right] \in \R$ is a standard deviation of all
components of $\tH\supp{p}$.
However, conversions such as
$\tH\supp{p} / \mathrm{Std}_\mathrm{component}\left[\tH\supp{p}\right]$
and
$\tH\supp{p} - \mathrm{Mean}\left[\tH\supp{p}\right]$
are not isometric transformation equivariant, assuming that
$\mathrm{Std}_\mathrm{component}\left[\tH\supp{p}\right] \in \R^{d^p}$
is a component-wise standard deviation.

\subsection{Scaling}
Because the concrete instance of IsoAM $\tilde{\tD}$ corresponds to the differential
operator, the scale of the output after operations regarding $\tilde{D}$ can be huge.
Thus, we rescale $\tilde{\tD}$ using the scaling factor
$\left[\mathrm{Mean}_{\mathrm{sample}, i}(
\tilde{\etD}_{ii;;1}^2
+ \tilde{\etD}_{ii;;2}^2
+ \tilde{\etD}_{ii;;3}^2
)\right]^{1/2}$,
where $\mathrm{Mean}_{\mathrm{sample}, i}$ denotes the mean over the samples and vertices.

\subsection{Implementation}
Because an adjacency matrix $\mA$ is usually a sparse matrix for a regular mesh,
$\mA(m)$ in equation~\ref{eq:d} is also a sparse matrix for a sufficiently small $m$.
Thus, we can leverage sparse matrix multiplication in the IsoGCN computation.
This is one major reason why IsoGCNs can compute rapidly.
If the multiplication (tensor product or contraction) of IsoAMs must be computed multiple times the 
associative property of the IsoAM can be utilized.

For instance, it is apparent that
$\left[ \bigotimes^k \tG \right] \ast \tH\supp{0} = \tG \otimes (\tG \otimes \dots (\tG \ast \tH\supp{0}))$.
Assuming that the number of nonzero elements in $\mA(m)$ equals $n$ and
$\tH\supp{0} \in \mathbb{R}^{\VV \times f}$, then
the computational complexity of the right-hand side is $\mathcal{O}(n \VV f d^k)$.
This is an exponential order regarding $d$.
However, $d$ and $k$ are usually
small numbers (typically $d =3$ and $k \leq 4$).
Therefore one can compute an IsoGCN layer
with a realistic spatial dimension $d$ and tensor rank $k$ fast and memory efficiently.
In our implementation, both a sparse matrix operation and associative property are
utilized to realize fast computation.

\section{Experiment details: differential operator dataset}\label{sec:differential}
\subsection{Model architectures}
\begin{figure}[bth]
  \centering
  \includegraphics[trim={8cm 25cm 8cm 0cm},clip,width=0.99\textwidth]
  {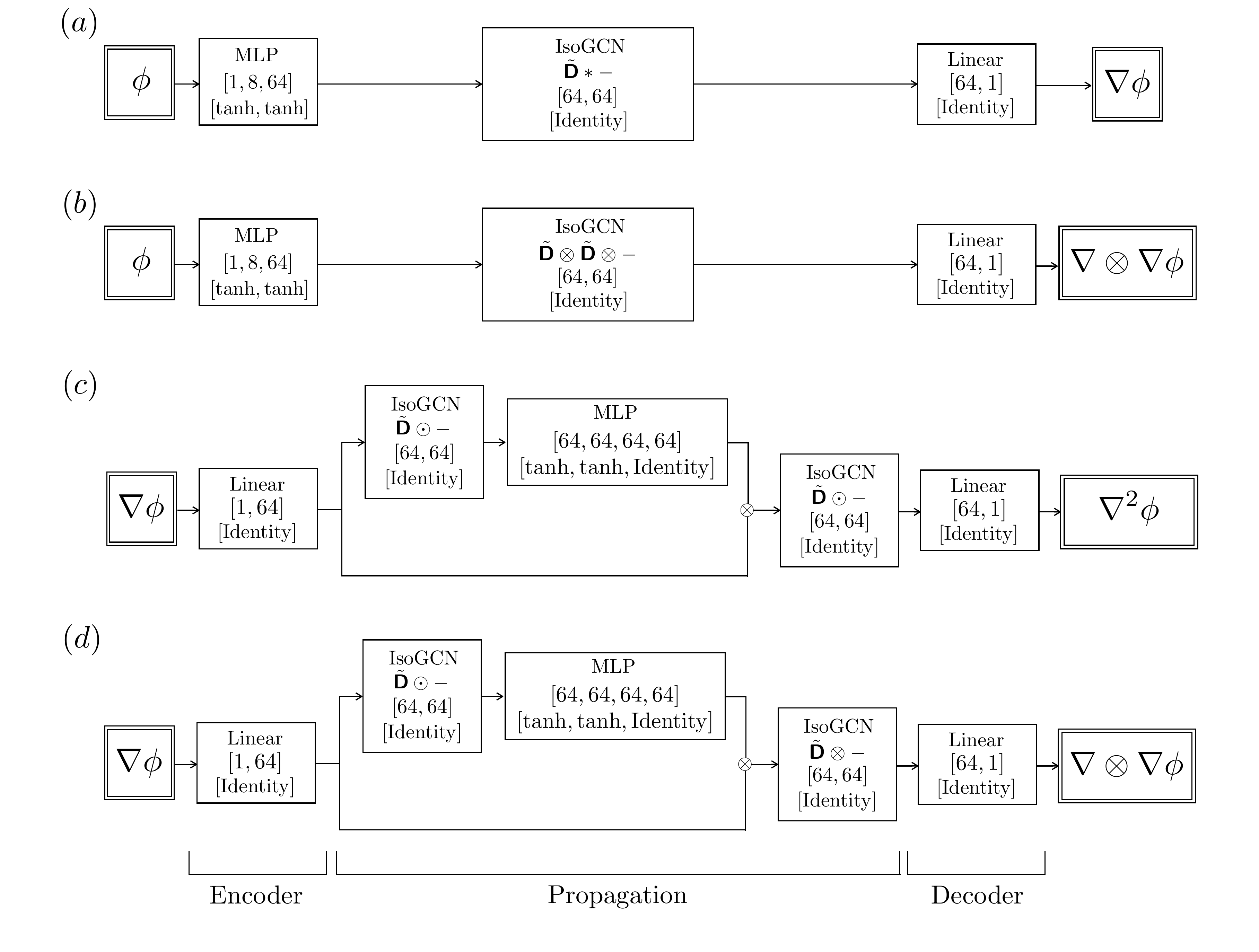}
  \caption{The IsoGCN model used for
  (a) the scalar field to the gradient field,
  (b) the scalar field to the Hessian field,
  (c) the gradient field to the Laplacian field,
  (d) the gradient field to the Hessian field of the gradient operator dataset.
  The numbers in each box denote the number of units.
  Below the unit numbers, the activation function used for each layer is also shown.
  $\otimes$ denotes the multiplication in the feature direction.
  }
  \label{fig:grid_networks}
\end{figure}

Figure~\ref{fig:grid_networks} represents the IsoGCN model used for the
differential operator dataset.
We used the $\tanh$ activation function as a nonlinear activation function
because we expect the target temperature field to be smooth.
Therefore, we avoid using non-differentiable
activation functions such as the rectified linear unit (ReLU)~\citep{nair2010rectified}.
For GCN and its variants, we simply replaced the IsoGCN layers with the corresponding ones.
We stacked $m \ (=2, 5)$ layers for GCN, GIN, GCNII, and Cluster-GCN.
We used an $m$ hop adjacency matrix for SGCN.

For the TFN and SE(3)-Transformer, we set the hyperparameters to have almost the same number of parameters
as in the IsoGCN model.
The settings of the hyperparameters are shown in Table~\ref{tab:diff_equiv_parameters}.

\bgroup
\def\arraystretch{1.0}
\begin{table}[tb]
  \centering
  \caption{Summary of the hyperparameter setting for both the TFN and SE(3)-Transformer.
  For the parameters not in the table, we used the default setting
  in the implementation of \url{https://github.com/FabianFuchsML/se3-transformer-public}.}
  \begin{tabular}{lcccc}
    & \boldsymbol{$0 \to 1$}
    & \boldsymbol{$0 \to 2$}
    & \boldsymbol{$1 \to 0$}
    & \boldsymbol{$1 \to 2$}
    \\
    \hhline{=====}
    \# hidden layers & 1 & 1 & 1 & 1
    \\
    \# NL layers in the self-interaction & 1 & 1 & 1 & 1
    \\
    \# channels & 24 & 20 & 24 & 24
    \\
    \# maximum rank of the hidden layers & 1 & 2 & 1 & 2
    \\
    \# nodes in the radial function & 16 & 8 & 16 & 22
  \end{tabular}
  \label{tab:diff_equiv_parameters}
\end{table}
\egroup

\subsection{Result details}
Table~\ref{tab:detailed_differential_dataset_results} represents the detailed comparison of training results.
The results show that an IsoGCN outperforms other GCN models for all settings.
Compared to other equivariant models, IsoGCN has competitive performance compared to equivariant models with shorter
inference time as shown in Table~\ref{tab:differential_speed}.
Therefore, it can be found out the proposed model has a strong expressive power
to express differential regarding space with less computation resources compared to
the TFN and SE(3)-Transformer.

\bgroup
\def\arraystretch{1.0}
\begin{table}[tb]
  \centering
  \caption{Summary of the test losses (mean squared error
  $\pm$ the standard error of the mean in the original scale)
  of the differential operator dataset:
  $0 \rightarrow 1$
  (the scalar field to the gradient field),
  $0 \rightarrow 2$
  (the scalar field to the Hessian field),
  $1 \rightarrow 0$
  (the gradient field to the Laplacian field),
  and $1 \rightarrow 2$
  (the gradient field to the Hessian field).
  Here, if ``$\vx$'' is ``Yes'', $\vx$ is also in the input feature.
  }

  \scalebox{0.9}{
  \begin{tabular}{lcccccc}
    \textbf{Method} & \textbf{\# hops} & \textbf{$\vx$}
    & \makecell{\textbf{Loss of }\boldsymbol{$0 \to 1$}\\$\times 10^{-5}$}
    & \makecell{\textbf{Loss of }\boldsymbol{$0 \to 2$}\\$\times 10^{-6}$}
    & \makecell{\textbf{Loss of }\boldsymbol{$1 \to 0$}\\$\times 10^{-6}$}
    & \makecell{\textbf{Loss of }\boldsymbol{$1 \to 2$}\\$\times 10^{-6}$}
    \\
    \hhline{=======}
    \hline
    \multirow{4}{*}{GIN}
    & 2 & No & 151.19 $\pm$ 0.53 & 49.10 $\pm$ 0.36 & 542.52 $\pm$ 2.14 & 59.65 $\pm$ 0.46
    \\
    & 2 & Yes & 147.10 $\pm$ 0.51 & 47.56 $\pm$ 0.35 & 463.79 $\pm$ 2.08 & 50.73 $\pm$ 0.40
    \\
    & 5 & No & 151.18 $\pm$ 0.53 & 48.99 $\pm$ 0.36 & 542.54 $\pm$ 2.14 & 59.64 $\pm$ 0.46
    \\
    & 5 & Yes & 147.07 $\pm$ 0.51 & 47.35 $\pm$ 0.35 & 404.92 $\pm$ 1.74 & 46.18 $\pm$ 0.39
    \\

    \hline
    \multirow{4}{*}{GCNII}
    & 2 & No & 151.18 $\pm$ 0.53 & 43.08 $\pm$ 0.31 & 542.74 $\pm$ 2.14 & 59.65 $\pm$ 0.46
    \\
    & 2 & Yes & 151.14 $\pm$ 0.53 & 40.72 $\pm$ 0.29 & 194.65 $\pm$ 1.00 & 45.43 $\pm$ 0.36
    \\
    & 5 & No & 151.11 $\pm$ 0.53 & 32.85 $\pm$ 0.23 & 542.65 $\pm$ 2.14 & 59.66 $\pm$ 0.46
    \\
    & 5 & Yes & 151.13 $\pm$ 0.53 & 31.87 $\pm$ 0.22 & 280.61 $\pm$ 1.30 & 39.38 $\pm$ 0.34
    \\

    \hline
    \multirow{4}{*}{SGCN}
    & 2 & No & 151.17 $\pm$ 0.53 & 50.26 $\pm$ 0.38 & 542.90 $\pm$ 2.14 & 59.65 $\pm$ 0.46
    \\
    & 2 & Yes & 151.12 $\pm$ 0.53 & 49.96 $\pm$ 0.37 & 353.29 $\pm$ 1.49 & 59.61 $\pm$ 0.46
    \\
    & 5 & No & 151.12 $\pm$ 0.53 & 55.02 $\pm$ 0.42 & 542.73 $\pm$ 2.14 & 59.64 $\pm$ 0.46
    \\
    & 5 & Yes & 151.16 $\pm$ 0.53 & 55.08 $\pm$ 0.42 & 127.21 $\pm$ 0.63 & 56.97 $\pm$ 0.44
    \\

    \hline
    \multirow{4}{*}{GCN}
    & 2 & No & 151.23 $\pm$ 0.53 & 49.59 $\pm$ 0.37 & 542.54 $\pm$ 2.14 & 59.64 $\pm$ 0.46
    \\
    & 2 & Yes & 151.14 $\pm$ 0.53 & 47.91 $\pm$ 0.35 & 542.68 $\pm$ 2.14 & 59.60 $\pm$ 0.46
    \\
    & 5 & No & 151.18 $\pm$ 0.53 & 50.58 $\pm$ 0.38 & 542.53 $\pm$ 2.14 & 59.64 $\pm$ 0.46
    \\
    & 5 & Yes & 151.14 $\pm$ 0.53 & 48.50 $\pm$ 0.35 & 542.30 $\pm$ 2.14 & 25.37 $\pm$ 0.28
    \\

    \hline
    \multirow{4}{*}{Cluster-GCN}
    & 2 & No & 151.19 $\pm$ 0.53 & 33.39 $\pm$ 0.24 & 542.54 $\pm$ 2.14 & 59.66 $\pm$ 0.46
    \\
    & 2 & Yes & 147.23 $\pm$ 0.51 & 32.29 $\pm$ 0.24 & 167.73 $\pm$ 0.83 & 17.72 $\pm$ 0.17
    \\
    & 5 & No & 151.15 $\pm$ 0.53 & 28.79 $\pm$ 0.21 & 542.51 $\pm$ 2.14 & 59.66 $\pm$ 0.46
    \\
    & 5 & Yes & 146.91 $\pm$ 0.51 & 26.60 $\pm$ 0.19 & 185.21 $\pm$ 0.99 & 18.18 $\pm$ 0.20
    \\

    \hline
    \multirow{2}{*}{TFN}
    & 2 & No & 2.47 $\pm$ 0.02 & OOM & 26.69 $\pm$ 0.24 & OOM
    \\
    & 5 & No & OOM & OOM & OOM & OOM
    \\

    \hline
    \multirow{2}{*}{SE(3)-Trans.}
    & 2 & No & \textbf{1.79} $\pm$ 0.02 & \textbf{3.50} $\pm$ 0.04 & \textbf{2.52} $\pm$ 0.02 & OOM
    \\
    & 5 & No & 2.12 $\pm$ 0.02 & OOM & 7.66 $\pm$ 0.05 & OOM
    \\

    \hline
    \multirow{2}{*}{\textbf{IsoGCN} (Ours)}
    & 2 & No & 2.67 $\pm$ 0.02 & 6.37 $\pm$ 0.07 & 7.18 $\pm$ 0.06 & \textbf{1.44} $\pm$ 0.02
    \\
    & 5 & No & 14.19 $\pm$ 0.10 & 21.72 $\pm$ 0.25 & 34.09 $\pm$ 0.19 & 8.32 $\pm$ 0.09
  \end{tabular}
  }
  \label{tab:detailed_differential_dataset_results}
\end{table}
\egroup

\bgroup
\def\arraystretch{1.0}
\begin{table}[tb]
  \centering
  \caption{Summary of the inference time on the test dataset.
  $0 \rightarrow 1$ corresponds to the scalar field to the gradient field,
  and $0 \rightarrow 2$ corresponds to the scalar field to the Hessian field.
  Each computation was run on the same GPU (NVIDIA Tesla V100 with 32 GiB memory).
  OOM denotes the out-of-memory of the GPU.}
  \begin{tabular}{lcccc}
    & \multicolumn{2}{c}{\boldsymbol{$0 \to 1$}}
    & \multicolumn{2}{c}{\boldsymbol{$0 \to 2$}}
    \\[3pt]
    \textbf{Method}
    & \textbf{\# parameters} & \textbf{Inference time [s]}
    & \textbf{\# parameters} & \textbf{Inference time [s]}
    \\
    \hhline{=====}
    TFN & 5264 & 3.8 & 5220 & OOM
    \\
    SE(3)-Trans. & 5392 & 4.0 & 5265 & 9.2
    \\
    \textbf{IsoGCN} (Ours) & 4816 & 0.4 & 4816 & 0.7
  \end{tabular}
  \label{tab:differential_speed}
\end{table}
\egroup

\section{Experiments details: anisotropic nonlinear heat equation dataset}\label{sec:nl_tensor}
\subsection{Dataset}\label{sec:dataset}
The purpose of the experiment was to solve the anisotropic nonlinear heat diffusion under
an adiabatic boundary condition.
The governing equation is defined as follows:
\begin{align}
  \Omega &\subset \R^3,
  \\
  \pdiff{T(\vx, t)}{t} &= \nabla \cdot \mC(T(\vx, t)) \nabla T(\vx, t), \text{in $\Omega$},
  \\
  T(\vx, t = 0) &= T_{0.0}(\vx), \text{in $\Omega$},
  \\
  \nabla T(\vx, t)|_{\vx = \vx_b} \cdot \vn(\vx_b) &= 0, \text{on $\partial \Omega$},
\end{align}
where
$T$ is the temperature field,
$T_{0.0}$ is the initial temperature field,
$\mC \in \R^{d \times d}$ is an anisotropic diffusion tensor
and $\vn(\vx_b)$ is the normal vector at $\vx_b \in \partial \Omega$.
Here, $\mC$ depends on temperature thus the equation is nonlinear.
We randomly generate $\mC(T = -1)$ for it to be a positive semidefinite symmetric tensor
with eigenvalues varying from 0.0 to 0.02.
Then, we defined the linear temperature dependency the slope of which is $- \mC(T = -1) / 4$.
The function of the anisotropic diffusion tensor is uniform for each sample.
The task is defined to predict the temperature field at $t = 0.2, 0.4, 0.6, 1.0$
($T_{0.2}, T_{0.4}, T_{0.6}, T_{0.8}, T_{1.0}$)
from the given initial temperature field, material property, and mesh geometry.
However, the performance is evaluated only with $T_{1.0}$ to focus on the predictive performance.
We inserted other output features to stabilize the trainings.
Accordingly, the diffusion number of this problem is
$\mC \Delta t / (\Delta x)^2 \simeq 10.0^{4}$
assuming $\Delta x \simeq 10.0^{-3}$.

\begin{figure}[bth]
  \centering
  \includegraphics[trim={0cm 0cm 0cm 0cm},clip,width=0.99\textwidth]
  {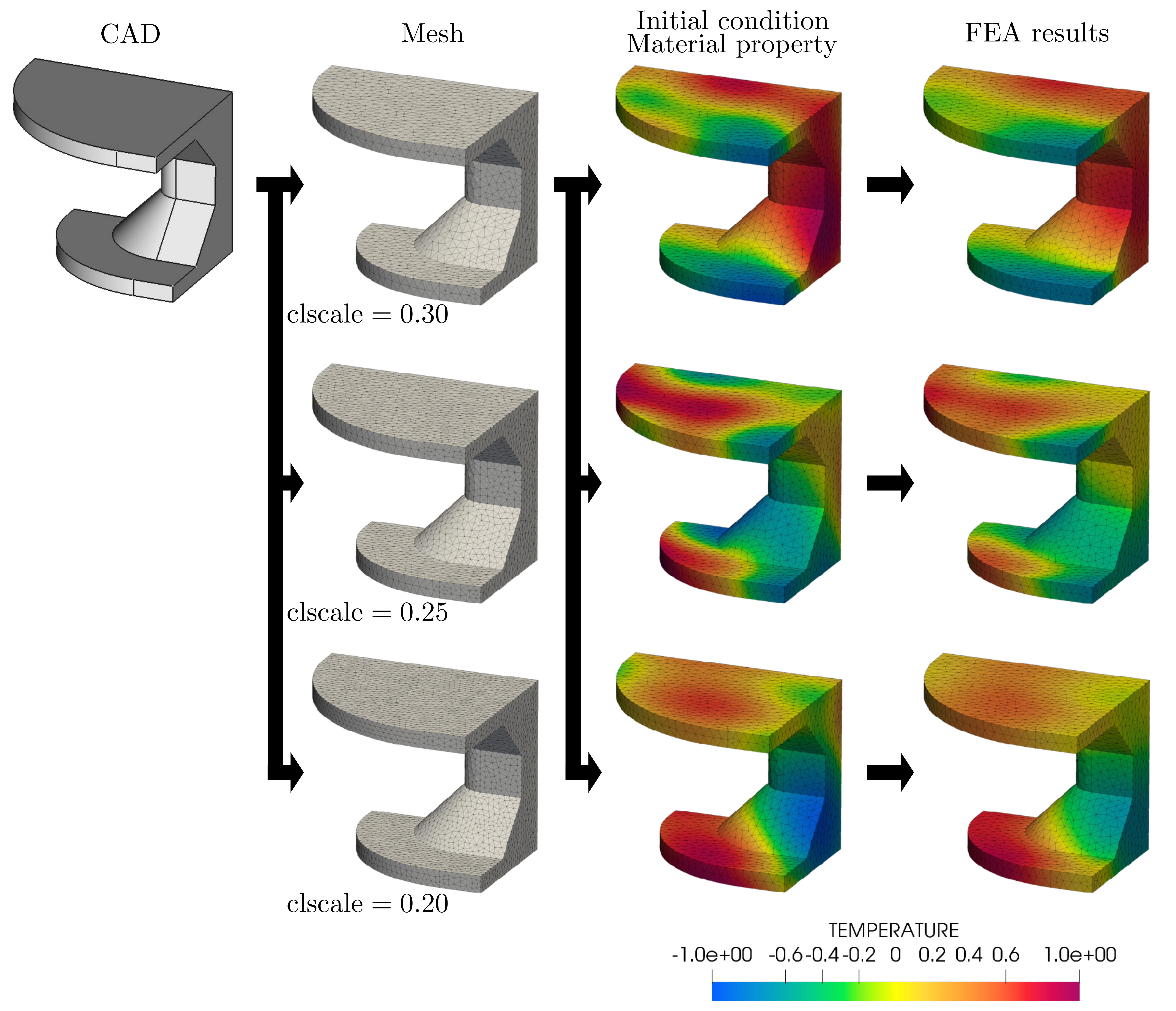}
  \caption{The process of generating the dataset.
  A smaller clscale parameter generates smaller meshes.
  }
  \label{fig:process}
\end{figure}

Figure~\ref{fig:process} represents the process of generating the dataset.
We generated up to 9 FEA results for each CAD shape.
To avoid data leakage in terms of the CAD shapes, we first split them
into training, validation, and test datasets, and then applied the following process.

Using one CAD shape, we generated up to three meshes using clscale
(a control parameter of the mesh characteristic lengths) $=$ 0.20, 0.25, and 0.30.
To facilitate the training process, we scaled the meshes to fit into a
cube with an edge length equal to 1.

Using one mesh, we generated three initial conditions randomly using a Fourier series of the 2nd to 10th orders.
We then applied an FEA to each initial condition and material property determined randomly as described above.
We applied an implicit method to solve time evolutions and a direct method
to solve the linear equations.
The FEA time step $\Delta t$ was set to 0.01.

During this process, some of the meshes or FEA results may not have been available due to
excessive computation time or non-convergence.
Therefore, the size of the dataset was not exactly equal to the number multiplied by 9.
Finally, we obtained 439 FEA results for the training dataset, 143 FEA results for the
validation dataset, and 140 FEA results for the test dataset.

\subsection{Input features}
To express the geometry information, we extracted the effective volume of
the $i$th vertex
$V\sup{effective}_i$ and
the mean volume of the $i$th vertex
$V\sup{mean}_i$, which are defined as follows:
\begin{align}
  V\sup{effective}_i &= \sum_{e \in \mathcal{N}^e_i} \frac{1}{4} V_e\label{eq:v_effective},
  \\
  V\sup{mean}_i &= \frac{\sum_{e \in \mathcal{N}^e_i} V_e}{\vert \mathcal{N}^e_i \vert}\label{eq:v_mean},
\end{align}
where
$\mathcal{N}^e_i$ is the set of elements, including the $i$th vertex.

For GCN or its variant models, we tested several combinations of input vertex features
$T_{0.0}$,
$\mC$,
$V\sup{effective}$,
$V\sup{mean}$, and
$\vx$ (Table~\ref{tab:detailed_results}).
For the IsoGCN model, inputs were
$T_{0.0}$,
$\mC$,
$V\sup{effective}$, and
$V\sup{mean}$.

\subsection{Model architectures}
\begin{figure}[bth]
  \centering
  \includegraphics[trim={4cm 37cm 3cm 0cm},clip,width=0.99\textwidth]
  {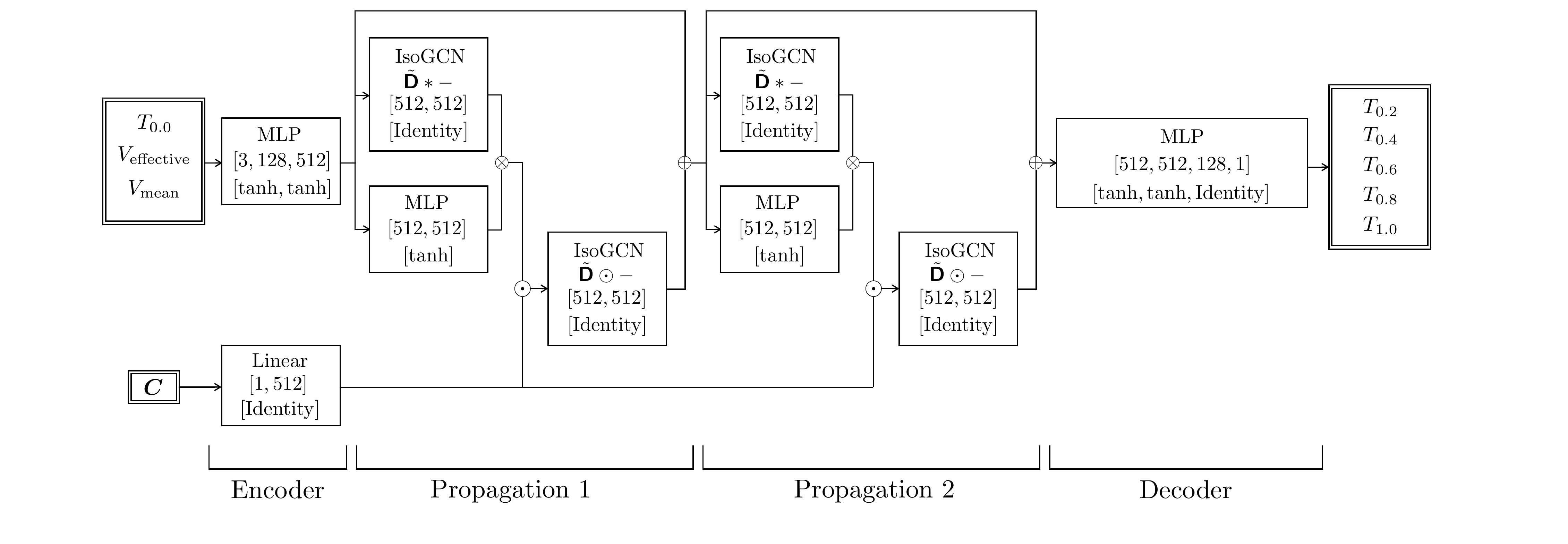}
  \caption{The IsoGCN model used for the anisotropic nonlinear heat equation dataset.
  The numbers in each box denote the number of units.
  Below the unit numbers, the activation function used for each layer is also shown.
  $\otimes$ denotes multiplication in the feature direction,
  $\odot$ denotes the contraction,
  and $\oplus$ denotes the addition in the feature direction.
  }
  \label{fig:nl_tensor_network}
\end{figure}

Figure~\ref{fig:nl_tensor_network} represents the IsoGCN model used for the
anisotropic nonlinear heat equation dataset.
We used the $\tanh$ activation function as a nonlinear activation function
because we expect the target temperature field to be smooth.
Therefore, we avoid using non-differentiable
activation functions such as the rectified linear unit (ReLU)~\citep{nair2010rectified}.
Although the model looks complicated, one propagation block corresponds to the
first-order Taylor expansion
$T(t+\Delta t) \simeq \nabla \mC \odot \nabla T(t) + T(t)$
because the propagation block is expressed as
$\tilde{\tD} \odot \mC \odot \mathrm{MLP}(T) \tilde{\tD} \ast T + T$,
where $T$ denotes the rank-0 tensor input to the propagation block.
By stacking this propagation block $p$ times, we can approximate the $p$th order Taylor expansion
of the anisotropic nonlinear heat equation.

For GCN and its variants, we simply replaced the IsoGCN layers with the corresponding ones.
We stacked $m \ (=2, 5)$ layers for GCN, GIN, GCNII, and Cluster-GCN.
We used an $m$ hop adjacency matrix for SGCN.

For the TFN and SE(3)-Transformer, we set the hyperparameters to as many parameters as possible that would
fit on the GPU because the TFN and SE(3)-Transformer with almost the same number
of parameters as in IsoGCN did not fit on the GPU we used (NVIDIA Tesla V100 with 32 GiB memory).
The settings of the hyperparameters are shown in Table~\ref{tab:nl_equiv_parameters}.

\bgroup
\def\arraystretch{1.0}
\begin{table}[tb]
  \centering
  \caption{Summary of the hyperparameter setting for both the TFN and SE(3)-Transformer.
  For the parameters not written in the table, we used the default setting
  in the implementation of \url{https://github.com/FabianFuchsML/se3-transformer-public}.}
  \begin{tabular}{lc}
    \# hidden layers & 1
    \\
    \# NL layers in the self-interaction & 1
    \\
    \# channels & 16
    \\
    \# maximum rank of the hidden layers & 2
    \\
    \# nodes in the radial function & 32
  \end{tabular}
  \label{tab:nl_equiv_parameters}
\end{table}
\egroup

\subsection{Result details}
\begin{figure}[tb]
  \centering
  \includegraphics[trim={2cm 10cm 2cm 0cm},clip,width=0.99\textwidth]
  {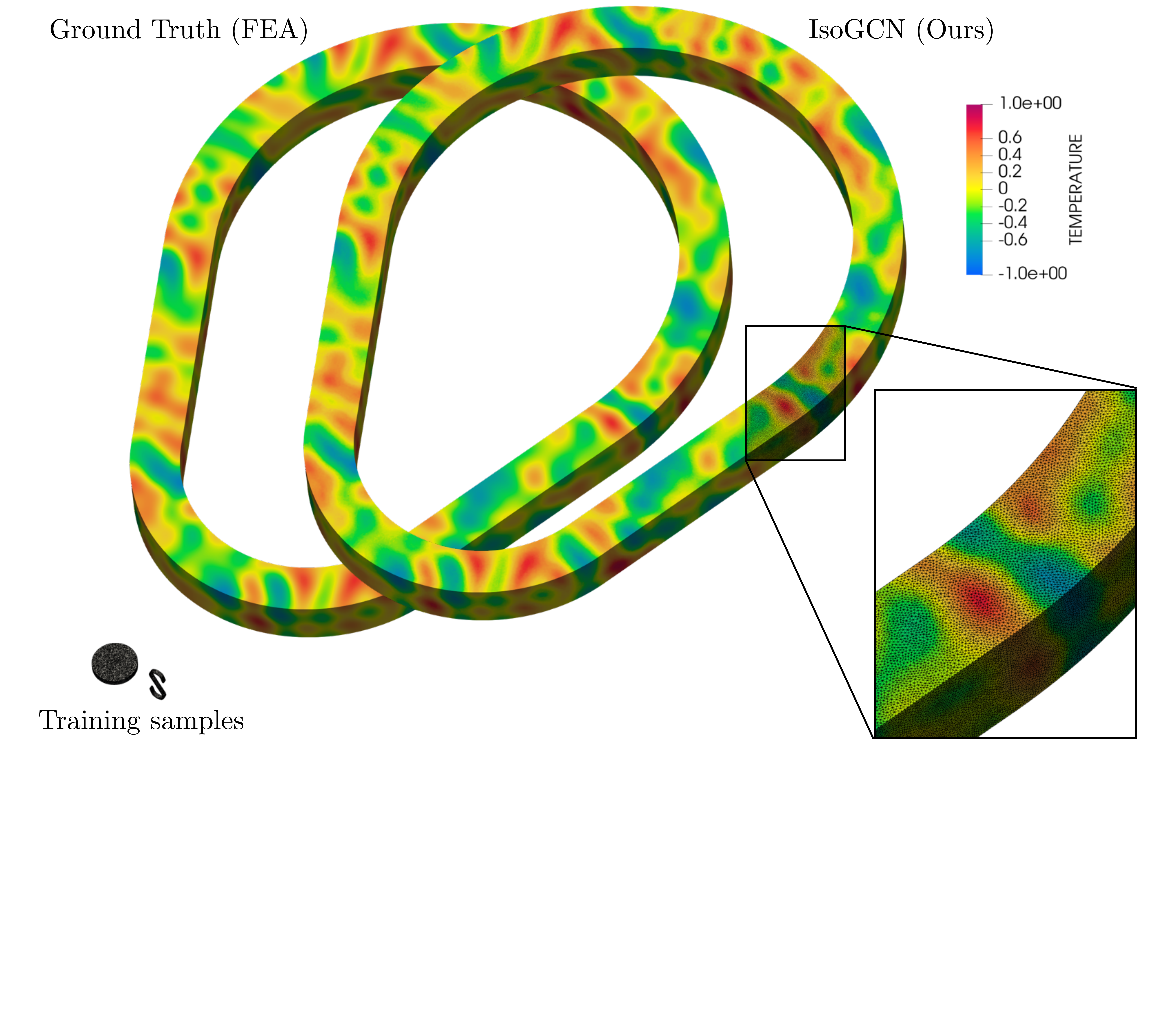}
  \caption{Comparison between (left) samples in the training dataset,
  (center) ground truth computed through FEA,
  and (right) IsoGCN inference result.
  For both the ground truth and inference result, $\VV = 1,011,301$.
  One can see that IsoGCN can predict the temperature field for a mesh,
  which is much larger than these in the training dataset.
  }
  \label{fig:larger}
\end{figure}

Table~\ref{tab:detailed_results} shows a detailed comparison of the training results.
The inclusion of $\vx$ in the input features of the baseline models did not improve the performance.
In addition, if $\vx$ is included in the input features,
a loss of the generalization capacity for larger shapes compared to the training dataset may result as it
extrapolates.
The proposed model achieved the best performance
compared to the baseline models considered.
Therefore, we concluded that the essential features regarding the mesh shapes are
included in $\tilde{\tD}$.
Besides, IsoGCN can scale up to meshes with 1M vertices as shown in Figure~\ref{fig:larger}.

\bgroup
\def\arraystretch{1.1}
\begin{table}[tb]
  \centering
  \caption{Summary of the test losses (mean squared error
  $\pm$ the standard error of the mean in the original scale) of the anisotropic nonlinear heat dataset.
  Here, if ``$\vx$'' is ``Yes'', $\vx$ is also in the input feature.
  OOM denotes the out-of-memory on the applied GPU (32 GiB).}
  \begin{tabular}{lccc}
    \textbf{Method} & \textbf{\# hops} & \boldsymbol{$\vx$}
    & \makecell{\textbf{Loss}\\$\times 10^{-3}$}
    \\
    \hhline{====}
    \multirow{4}{*}{GIN}
    & 2 & No & 16.921 $\pm$ 0.040 %
    \\
    & 2 & Yes & 18.483 $\pm$ 0.025 %
    \\
    & 5 & No & 22.961 $\pm$ 0.056 %
    \\
    & 5 & Yes & 17.637 $\pm$ 0.046 %
    \\
    \hline
    \multirow{4}{*}{GCN}
    & 2 & No & 10.427 $\pm$ 0.028 %
    \\
    & 2 & Yes & 11.610 $\pm$ 0.032 %
    \\
    & 5 & No & 12.139 $\pm$ 0.031 %
    \\
    & 5 & Yes & 11.404 $\pm$ 0.032 %
    \\
    \hline
    \multirow{4}{*}{GCNII}
    & 2 & No & 9.595 $\pm$ 0.026 %
    \\
    & 2 & Yes & 9.789 $\pm$ 0.028 %
    \\
    & 5 & No & 8.377 $\pm$ 0.024 %
    \\
    & 5 & Yes & 9.172 $\pm$ 0.028 %
    \\
    \hline
    \multirow{4}{*}{Cluster-GCN}
    & 2 & No & 7.266 $\pm$ 0.021 %
    \\
    & 2 & Yes & 8.532 $\pm$ 0.023 %
    \\
    & 5 & No & 8.680 $\pm$ 0.024 %
    \\
    & 5 & Yes & 10.712 $\pm$ 0.030 %
    \\
    \hline
    \multirow{4}{*}{SGCN}
    & 2 & No & 7.317 $\pm$ 0.021 %
    \\
    & 2 & Yes & 9.083 $\pm$ 0.026 %
    \\
    & 5 & No & 6.426 $\pm$ 0.018 %
    \\
    & 5 & Yes & 6.519 $\pm$ 0.020 %
    \\
    \hline
    \multirow{2}{*}{TFN}
    & 2 & No & 15.661 $\pm$ 0.019 %
    \\
    & 5 & No & OOM
    \\
    \multirow{2}{*}{SE(3)-Trans.}
    & 2 & No & 14.164 $\pm$ 0.018 %
    \\
    & 5 & No & OOM
    \\
    \hline
    \multirow{2}{*}{\textbf{IsoGCN} (Ours)}
    & 2 & No & 4.674 $\pm$ 0.014 %
    \\
    & 5 & No & \textbf{2.470} $\pm$ 0.008 %
  \end{tabular}
  \label{tab:detailed_results}
\end{table}
\egroup
\end{document}